\pdfoutput=1
\documentclass[mnsc,nonblindrev]{informs3}

\DoubleSpacedXI %
\usepackage{endnotes}
\let\footnote=\endnote
\let\enotesize=\normalsize
\def\notesname{Endnotes}%
\def\makeenmark{\hbox to1.275em{\theenmark.\enskip\hss}}
\def\enoteformat{\rightskip0pt\leftskip0pt\parindent=1.275em
  \leavevmode\llap{\makeenmark}}

\usepackage{natbib}
 \bibpunct[, ]{(}{)}{,}{a}{}{,}%
 \def\bibfont{\small}%

\usepackage{mathrsfs}
\usepackage{amsmath}
\usepackage{amssymb}
\usepackage{float}
\usepackage{mathtools}
\usepackage{hyperref}
\usepackage{graphicx}
\usepackage{subcaption}
\usepackage{amsfonts}
\usepackage{xcolor}
 \usepackage{eurosym}
\usepackage{hyperref}
\hypersetup{colorlinks=true,breaklinks=true}
\usepackage{enumitem}
\usepackage[capitalise]{cleveref}

\usepackage[colorinlistoftodos]{todonotes}
\usepackage[short]{optidef}
\usepackage{amsmath}

\newenvironment{talign*}
 {\let\displaystyle\textstyle\csname align*\endcsname}
 {\endalign}

\TheoremsNumberedThrough     
\ECRepeatTheorems
\EquationsNumberedThrough    

\begin{document}

\RUNAUTHOR{Elmachtoub, Liang, McNellis}

\RUNTITLE{Decision Trees for Decision-Making under the Predict-then-Optimize Framework}


\TITLE{Decision Trees for Decision-Making under the Predict-then-Optimize Framework}

\ARTICLEAUTHORS{%
\AUTHOR{Adam N. Elmachtoub}
\AFF{Department of Industrial Engineering and Operations Research and Data Science Institute, Columbia University, NY, USA \EMAIL{adam@ieor.columbia.edu}} 
\AUTHOR{Jason Cheuk Nam Liang}
\AFF{Operations Research Center, Massachusetts Institute of Technology, MA, USA \\ \EMAIL{jcnliang@mit.edu}}
\AUTHOR{Ryan McNellis}
\AFF{Department of Industrial Engineering and Operations Research and Data Science Institute, Columbia University, NY, USA \\ Amazon\footnote{Publication written prior to Amazon employment}, NY, USA \\ \EMAIL{rtm2130@columbia.edu, rmcnell@amazon.com}}
} 

\ABSTRACT{%
We consider the use of decision trees for decision-making problems under the predict-then-optimize framework. That is, we would like to first use a decision tree to predict unknown input parameters of an optimization problem, and then make decisions by solving the optimization problem using the predicted parameters. A natural loss function in this framework is to measure the suboptimality of the decisions induced by the predicted input parameters, as opposed to measuring loss using input parameter prediction error. This natural loss function is known in the literature as the Smart Predict-then-Optimize (SPO) loss, and we propose a tractable methodology called SPO Trees (SPOTs) for training decision trees under this loss. SPOTs benefit from the interpretability of decision trees, providing an interpretable segmentation of contextual features into groups with distinct optimal solutions to the optimization problem of interest. We conduct several numerical experiments on synthetic and real data including the prediction of travel times for shortest path problems and predicting click probabilities for news article recommendation. We demonstrate on these datasets that SPOTs simultaneously provide higher quality decisions and significantly lower model complexity than other machine learning approaches (e.g., CART) trained to minimize prediction error.
}


\maketitle

\pdfoutput=1
\section{Introduction} \label{intro} 
Many decision-making problems of interest to practitioners can be framed as optimization problems containing uncertain input parameters to be estimated from data. For example, personalized advertising requires estimation of click/conversion probabilities as a function of user features, portfolio optimization problems necessitate accurate predictions of asset returns, and delivery routing problems require forecasts of travel times. A convenient and widely-utilized framework for addressing these problems is the \textit{predict-then-optimize} framework. 
Predict-then-optimize is a two step approach which \textit{(i)} first predicts any uncertain input parameters using a machine learning (ML) model trained on historical data, and \textit{(ii)} then generates decisions by solving the corresponding optimization problem using the predicted parameters. Typically, the ML models in this framework are trained using loss functions measuring prediction error (e.g., mean squared error) without considering the  impact of the predictions on the downstream optimization problem. However, for many practitioners, the primary interest is in obtaining near-optimal decisions \textit{from} the input parameter estimates rather than minimizing prediction error. In this work, we provide a methodology for training decision trees, under the predict-then-optimize framework, to minimize decision error rather than prediction error. 

A natural idea is to integrate the prediction task with the optimization task, training the ML models using a loss function which directly measures the suboptimality of the decisions induced by the predicted input parameters. \citet{elmachtoub2017smart} propose such a loss function for a broad class of decision-making problems, which they refer to as the \textit{Smart Predict-then-Optimize loss} (SPO loss). 
However, the authors note that training ML models using SPO loss is likely infeasible due to the SPO loss function being nonconvex and discontinuous (and therefore not differentiable). The authors therefore propose a convex surrogate loss function they refer to as SPO+ loss, which they show is Fisher consistent with respect to SPO loss under some assumptions. \citet{wilder2019melding} also note the nondifferentiability of SPO loss and modify the objective function of the nominal optimization problem to derive a differentiable, surrogate loss function. 
Both works demonstrate empirically that training ML models using the surrogate loss functions yields better decisions than models trained to minimize prediction error. However, the surrogate loss functions are not guaranteed to recover optimal decisions with respect to SPO loss and merely serve as approximations for computational feasibility. A practical and general methodology for training ML models using SPO loss directly has not yet been proposed.


In this work, we present algorithms for training decision trees to minimize SPO loss, which we call \textit{SPO Trees} (SPOTs). Despite the nonconvexity and discontunity of the SPO loss function, we show that the optimization problem for training decision tree models with respect to SPO loss can be greatly simplified through exploiting certain structural properties of decision trees. Therefore, to the best of our knowledge, we provide the first tractable methodology for training an ML model using SPO loss for a general class of decision-making problems. Decision trees are typically trained using ``greedy'' recursive partitioning approaches to minimize prediction error such as the popular CART algorithm \citep{breiman1984classification}; several recent works have also proposed integer programming strategies for training decision trees to optimality \citep{bertsimas2017optimal,gunluk2018optimal,verwer2019learning,hu2019optimal,aghaei2020learning}. We propose tractable extensions of the greedy and integer programming methodologies from the literature to train decision trees using SPO loss. We also provide methodology for training an \textit{ensemble} of SPO Trees to boost decision performance, which we refer to as \textit{SPO Forests}.  We conduct several numerical experiments on synthetic and real data demonstrating that SPOTs simultaneously find higher quality decisions while exhibiting significantly lower model complexity (i.e., depth) than other tree-building approaches trained to only minimize prediction error (e.g., CART). Implementations of our algorithms and experiments may be found at \href{https://github.com/rtm2130/SPOTree}{https://github.com/rtm2130/SPOTree}.

We remark that the use of decision trees for decision-making problems has seen increased attention in practice and recent literature due to their \textit{interpretability} \citep{kallus2017recursive,elmachtoub2017practical,ciocan2018interpretable, bertsimas2019optimal, aghaei2019learning, aouad2019market}. Decision trees for decision-making are seen as interpretable since their splits which map features to decisions are easily visualized. One of our key findings is that SPOTs end up being even more interpretable than trees trained to minimize prediction error as they require significantly less leaves to yield high-quality decisions. Finally, we note that decision trees exhibit several desirable properties as estimators. Namely, they are nonparametric, allowing them to capture nonlinear relationships and interaction terms which would have to be manually specified in other models such as linear regression.

\subsection{Literature Review}
There have been several approaches proposed in the recent literature for training decision tree models for optimal decision-making. \citet{bertsimas2019predictive} show how to properly leverage ML algorithms, including decision trees, in order to yield asymptotically optimal decisions to a class of stochastic optimization problems. However, their decision trees are trained in the same procedure as CART (but applied differently) and thus do not take into consideration the structure of the underlying decision-making problem. There has also been several recent works  on training decision trees for personalizing treatments among a finite set of possible options. \citet{kallus2017recursive} uses a loss function for training their trees which maximizes the efficacy of the recommended treatments rather than minimizing prediction error. 
\citet{bertsimas2019optimal} consider a similar treatment recommendation problem, but their approach uses an objective function involving a weighted combination of prediction and decision error. Our approach considers a more general class of decision-making problems potentially involving a large number of decisions represented by a general feasible region. \citet{aghaei2019learning} propose methodology for training decision trees for decision-making problems using a loss function which penalizes predictions that discriminate on sensitive features such as race or gender. However, their loss function does not consider the impact of predictions on downstream decisions, instead seeking to minimize prediction error. 

We also summarize a few additional approaches proposed in the literature which successfully apply other types of ML models to decision-making problems. \citet{kao2009directed} propose a loss function for training linear regression models which minimizes a convex combination between the prediction error and decision error. In addition to not considering decision tree models, their setting considers only quadratic optimization problems with no constraints. \citet{donti2017task} provide a more general methodology related to this line of work that relies on differentiating the optimization problem. 
\citet{wilder2019end} consider the problem of optimizing a function whose input is a graph structure that is unknown but can be estimated through prediction. Their end-to-end learning procedure involves constructing a simpler optimization problem in continuous space as a differentiable proxy for the more complex graph optimization problem.
\citet{wilder2019melding, mandi2019smart} consider training ML models using ``decision-focused'' loss functions for various combinatorial optimization problems; their methods do not attempt to minimize SPO loss directly but rather employ simpler surrogate loss functions. 
\citet{demirovic2019predict+} propose methodology for training linear regression models to directly minimize SPO loss, but their approach is specialized for ranking optimization problems. By contrast, we propose methodology for training decision trees under SPO loss for a more general class of optimization problems (which subsumes ranking problems as a special case). 



\pdfoutput=1
\section{The Predict-then-Optimize Framework}
In this section, we summarize the predict-then-optimize framework and the SPO loss proposed in \citet{elmachtoub2017smart}. We focus on a general class of decision-making problems which can be described by an optimization problem with known constraints and an unknown linear objective function (at the time of solving) which can be predicted from feature data. Many relevant problems of interest fall under this general structure,  include predicting travel times for shortest path problems, predicting demand for inventory management problems, and predicting returns for portfolio optimization.

We let  $S \subseteq \mathbb{R}^d$ denote the feasible region for the decisions, where $d$ is the dimension of the decision space. The decision-making problem can then defined mathematically as  
$z^*(c) = \min_{w \in S} c^Tw$, where $c \in \mathbb{R}^d$ is a cost vector of the optimization problem and $w \in \mathbb{R}^d$ is the vector of decision variables. Let $W^*(c)= \argmin_{w\in S} \{c^Tw\}$ denote the set of optimal decisions corresponding to $z^*(c)$, and let $w^*(c)$ denote an arbitrary individual member of the set $W^*(c)$. It is assumed that $S$ is specified in such a way that the computation of $w^*(c)$ and $z^*(c)$ are tractable for any cost vector $c$; for example, commercial optimization solvers are known to capably solve optimization problems with linear, conic, and/or  integer constraints. 

In the predict-then-optimize framework, the true cost vector is not known at the time of solving $w^*(\cdot)$ for an optimal decision, and thus a predicted cost vector $\hat{c}$ is used instead. Our predictions will rely on training a ML model from a given dataset $\{(x_1,c_1),(x_2,c_2),...,(x_n,c_n)\}$, where $x \in \mathbb{R}^p$ denote a vector of $p$ features available for predicting $c$. The $n$ feature-cost samples in the dataset are assumed to be independently and identically distributed according to an unknown joint distribution on $x$ and $c$. 
Let $\mathcal{H}$ denote a hypothesis class of candidate ML models $f: \mathbb{R}^p \rightarrow \mathbb{R}^d$ for predicting cost vectors from feature vectors, where $\hat{c} = f(x)$ is interpreted as the predicted cost vector associated with feature vector $x$ for model $f$. Finally, let $\ell(\cdot,\cdot): \mathbb{R}^d \times \mathbb{R}^d \rightarrow \mathbb{R}_+$ denote the loss function used to train the ML models, where $\ell(\hat{c},c)$ scores the loss incurred by a prediction of $\hat{c}$ when the true cost vector is $c$. Given a specified hypothesis class $\mathcal{H}$ and loss function $\ell(\cdot,\cdot)$, the ML models are trained through solving the following empirical risk minimization problem: 
\begin{align} \label{eqn:traingen}
f^* = \argmin_{f\in \mathcal{H}} \quad \frac{1}{n} \sum_{i = 1}^n \ell(f(x_i), c_i)
\end{align}
In words, the trained ML model $f^*$ is the model in the hypothesis class $\mathcal{H}$ which achieves the smallest average loss on the training data with respect to the given loss function $\ell(\cdot,\cdot)$. When presented with a new feature vector $x$, the model $f^*$ can be applied in predicting a cost vector $\hat{c} = f^*(x)$, and an optimal decision $w^*(\hat{c})$ is then proposed using the prediction $\hat{c}$.

One common loss function is mean squared error (MSE) loss, defined as $\ell_{MSE}(\hat{c}, c) := ||\hat{c}-c||_2^2$. 
By comparison, SPO loss scores predicted costs \textit{not} by their prediction error but rather by the quality of the decisions that they induce. Mathematically, SPO loss measures the excess cost $c^T w^*(\hat{c}) - z^*(c)$ incurred from making the (potentially) sub-optimal decision $w^*(\hat{c})$ implied by prediction $\hat{c}$ when the true cost is $c$.  Note that $W^*(\hat{c})$ may contain more than one optimal solution associated with $\hat{c}$. Therefore, \citet{elmachtoub2017smart} define SPO loss with respect to the \textit{worst-case} decision from a predicted cost vector $\hat{c}$, defined mathematically below:
\begin{align} \label{eqn:spoloss}
\ell_{SPO}(\hat{c}, c) := \max_{w \in W^*(\hat{c})} \{c^T w\} - z^*(c)\,.
\end{align}
The authors note that training ML models under SPO loss directly is likely infeasible, as SPO loss is nonconvex and discontinuous (and thus not differentiable) with respect to a given prediction $\hat{c}$. Therefore, the authors instead provide an algorithm for training linear models using a convex surrogate loss function called SPO+ loss.
\citet{wilder2019melding} also note the nondifferentiability of SPO loss and modify the objective function of the nominal optimization problem to derive a differentiable, surrogate loss function. In contrast to prior work, we provide multiple strategies for training decision trees using the SPO loss function \textit{directly}. Our methodology is presented in Section \ref{sec:spotmethods}.



\pdfoutput=1

\section{Decision Trees for Decision-Making}
\begin{figure}
	\centering
	\includegraphics[width=0.7\linewidth]{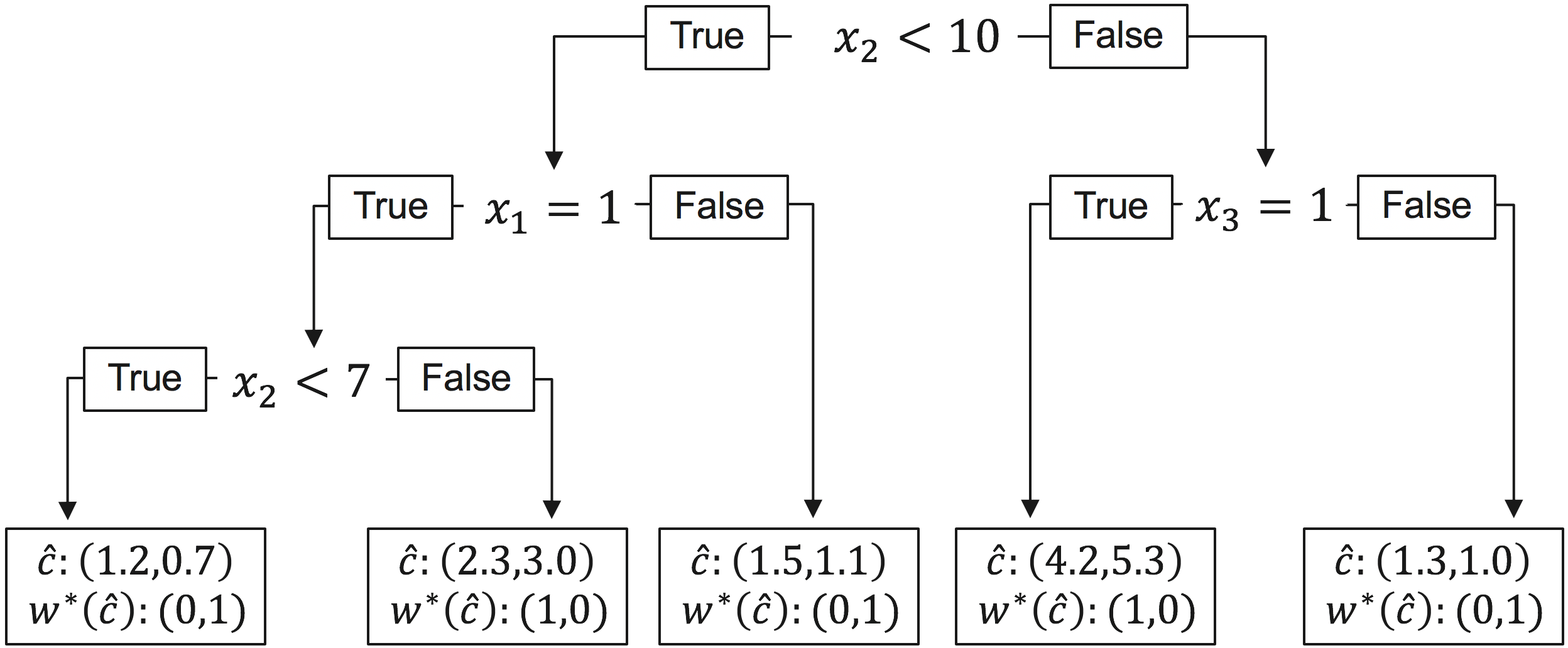}
	\caption{Decision tree for a shortest path problem with two edges.}
	\label{fig:spotEx}
\end{figure}

In this work, we utilize decision trees under the predict-then-optimize framework. To illustrate this concept, we consider a simple shortest path problem in a graph with two nodes and two candidate roads between them, each with unknown travel times (edge costs) $c_1$ and $c_2$.  We assume that there are $p = 3$ features available for predicting edge costs: $x_1$ is a binary feature to indicate a weekday, $x_2$ is the current hour of the day, and $x_3$ is a binary feature to indicate snowfall. The goal is to choose the path with the smallest cost given the observed features. An example of a decision tree applied to this problem is provided in Figure \ref{fig:spotEx}, although we note the same logic applies to an arbitrarily sized shortest path graph. Decision trees partition the feature space $\mathbb{R}^p$ through successive splits on components of the feature vector $x$. Each split takes the form of a yes-or-no question with respect to a single component. Continuous or ordinal features are split using inequalities, and categorical features are split using equalities. The partitions of $\mathbb{R}^p$ resulting from the decision tree splits are referred to as the \textit{leaves} of the tree. Each leaf assigns a single predicted cost vector $\hat{c}$ and associated decision $w^*(\hat{c})$ to all feature vectors which map to that leaf. We define the \textit{depth} of a leaf as the number of splits taken to reach that leaf. The depth of the tree is defined as the maximum of the depths of its leaves.

Decision trees are widely regarded as being very interpretable machine learning models, as the mapping from features to costs/decisions may be easily visualized and analyzed for insights. For example, in the decision tree of Figure \ref{fig:spotEx}, the second leaf from the left corresponds to the splits $x_2 < 10$, $x_1 = 1$, and $x_2 \geq 7$, which may be interpreted as the tree determining whether it is currently morning rush hour (i.e., a weekday between 7am and 10am). 

\subsection{An Illustrative Example} \label{sec:illustrative}

We provide a simple example to illustrate the behavior of decision trees trained using SPO loss versus MSE loss (i.e., SPOTs versus CARTs). We again consider the two edge shortest path problem from before, although we now assume there is only a single continuous feature $x$ available for predicting the travel times of the two edges. 
We generate a dataset of 10000 feature-cost pairs by (1) sampling 10000 feature values from a Uniform(0,1) distribution, and (2) computing each feature's associated edge cost by the equations $c_1 = 5x + 1.9$ and $c_2 = (5x+0.4)^2$ with no noise for the sake of illustration. We then train a decision tree to minimize SPO loss on this dataset, employing the SPOT training methods detailed in the next section. For sake of comparison, we also train a CART decision tree on the same dataset. CARTs are trained to minimize prediction error, specifically, mean-squared error in our experiments.

\begin{figure*}
	\centering
	\begin{subfigure}{0.48\linewidth}
		\centering
		\includegraphics[width=0.97\linewidth]{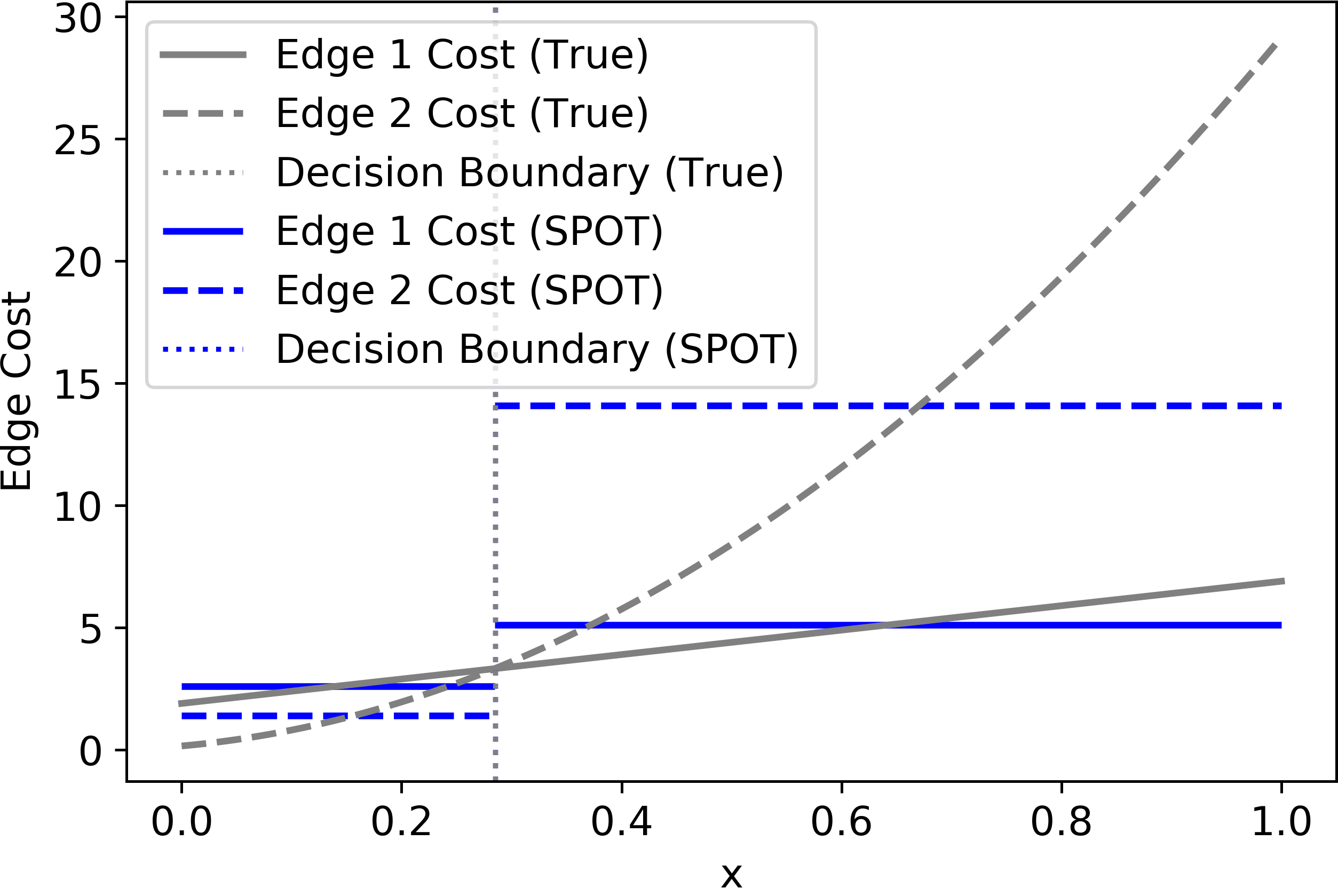}
		\caption{\small SPOT (Depth 1)}
		\vspace{0.1cm}
		\label{fig:casestudySPOT}
	\end{subfigure}%
	\begin{subfigure}{0.48\linewidth}
		\centering
		\includegraphics[width=0.97\linewidth]{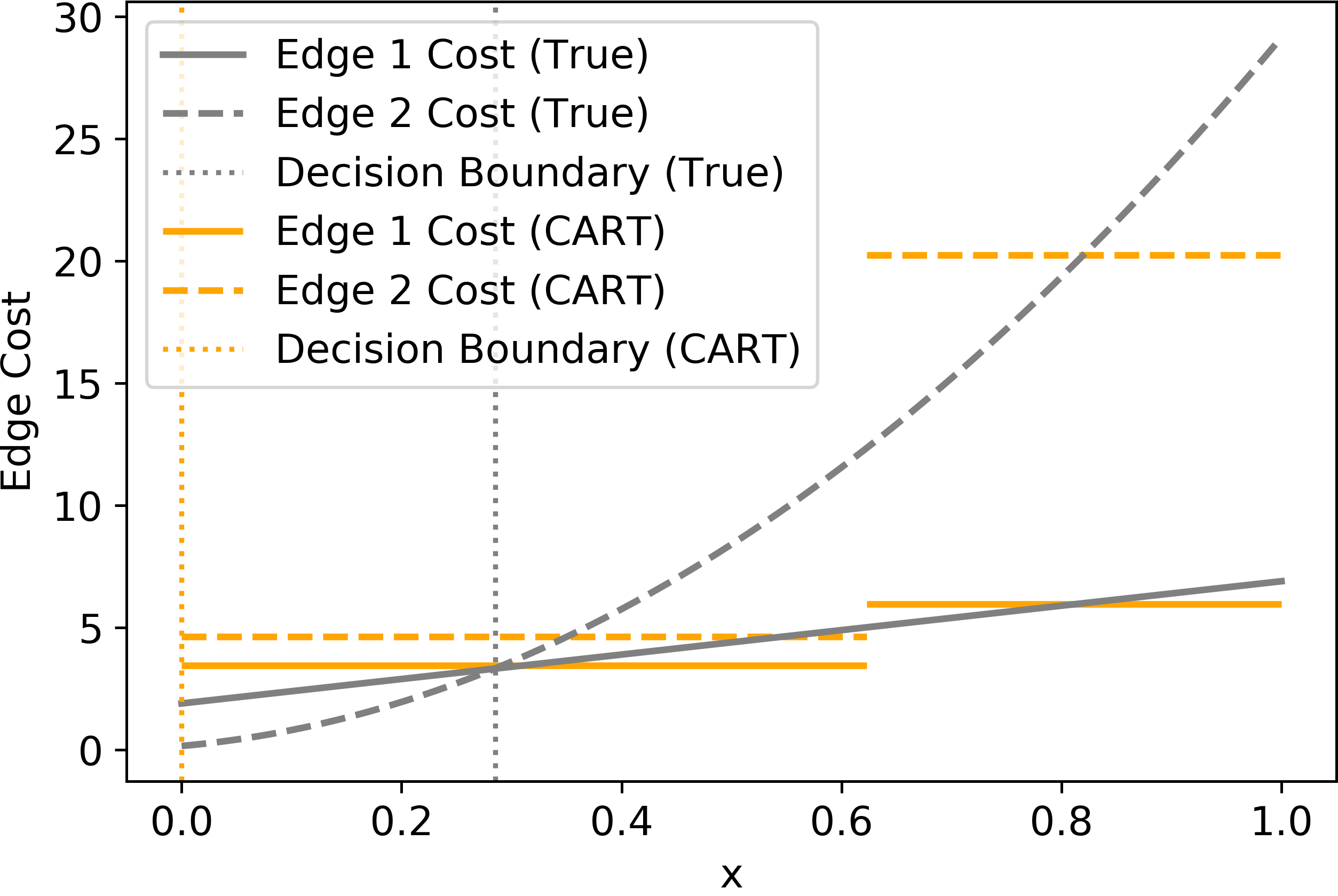}
		\caption{\small CART (Depth 1)}
		\vspace{0.1cm}
		\label{fig:casestudyCART1}
	\end{subfigure}\\
	\begin{subfigure}{0.48\linewidth}
		\centering
		\includegraphics[width=0.97\linewidth]{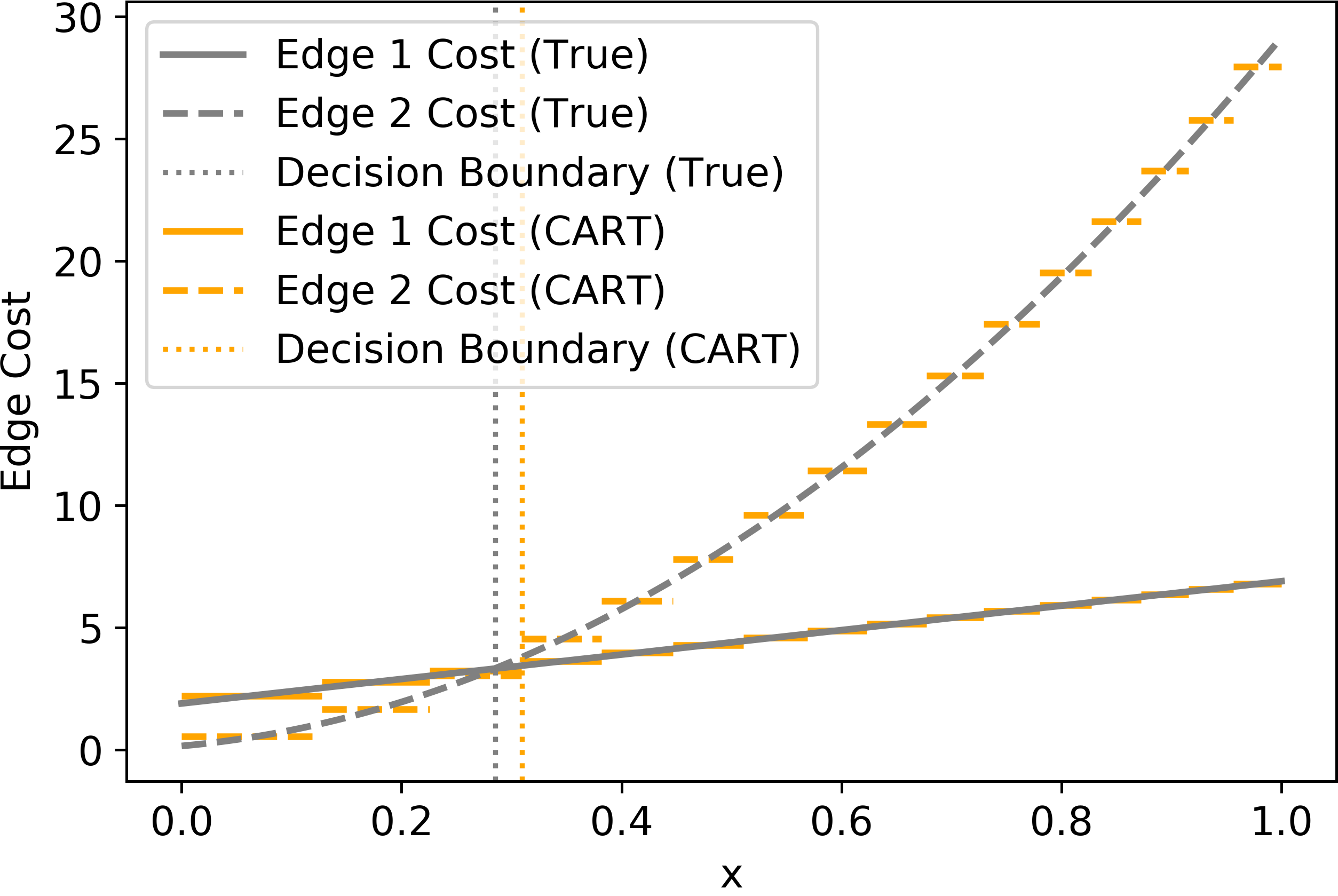}
		\caption{\small CART (Depth 4)}
		\vspace{0.1cm}
		\label{fig:casestudyCART4}
	\end{subfigure}%
	\begin{subfigure}{0.48\linewidth}
		\centering
		\includegraphics[width=0.97\linewidth]{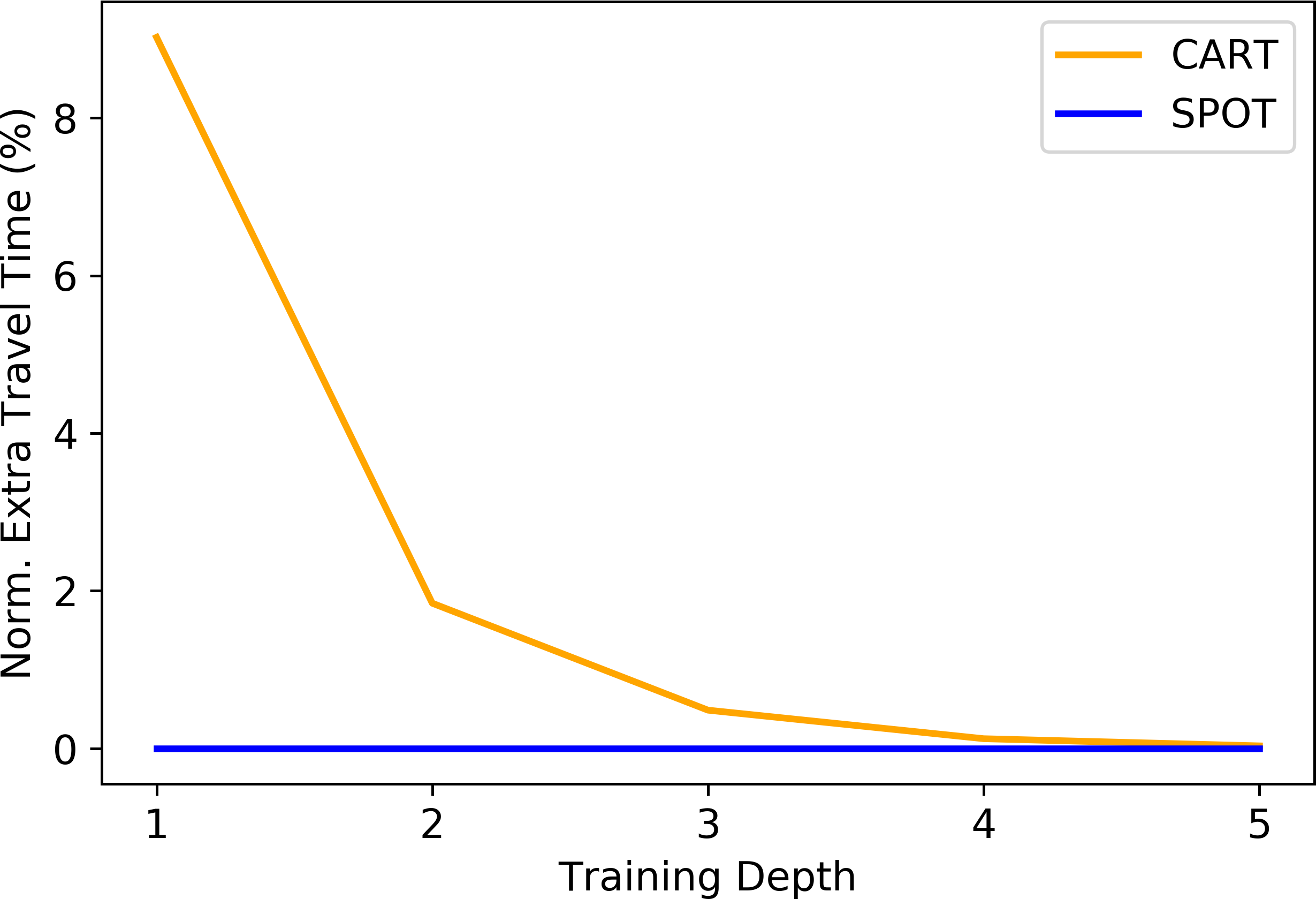}
		\caption{\small SPOT vs CART Loss}
		\vspace{0.1cm}
		\label{fig:casestudyCARTerrors}
	\end{subfigure}
	\caption{Predictive and decision performance of SPOT and CART decision trees. Figures (a)-(c) visualize the cost predictions of SPOT (blue) and CART (orange) alongside the true cost values (grey). Figure (d) plots the normalized extra travel time of the algorithms as a function of their trained tree depth.}
	\label{fig:casestudy}
\end{figure*}

The predictive and decision performance of the SPOT and CART training algorithms are given in Figure \ref{fig:casestudy}. Figures \ref{fig:casestudySPOT}-\ref{fig:casestudyCART4} visualize the cost predictions of the SPOT and CART algorithms and compare them against the true unknown edge costs. The two edge costs are equal at $x = 0.28$, at which point the optimal decision switches from taking edge 2 to taking edge 1. We therefore refer to the point $x = 0.28$ as the optimal or true decision boundary, and is referenced in the figures as a grey vertical line. We also include in the figures the decision boundaries implied by the cost predictions of the SPOT and CART algorithms. 

As shown in Figure \ref{fig:casestudySPOT}, the SPO Tree immediately identifies the correct decision boundary through the split ``$x < 0.28$''. This behavior is unsurprising, as any other individual split would have resulted in a suboptimal SPO loss incurred on the training set. Each leaf of the SPO tree yields a single predicted cost vector, which is visualized by the flat prediction lines in the regions ``$x < 0.28$'' and ``$x \geq 0.28$'' of the figure.

Figures \ref{fig:casestudyCART1} and \ref{fig:casestudyCART4} show the cost vector predictions of the CART algorithm. When trained to a depth of 1 (i.e., a single split), CART results in a severely incorrect decision boundary at $x=0$. This occurs because CART splits at $x < 0.62$, and in each of the resulting leaves from this split edge 2 is predicted to have a higher cost than edge 1. Therefore the CART algorithm incorrectly predicts that path 1 is always optimal, resulting in the decision boundary of $x = 0$. The CART algorithm does not split on the optimal decision boundary because this is \textit{not} the split which minimizes cost prediction error on the training set. Consequently, although the cost 
\textit{predictions} of CART may be more accurate, the implied shortest path \textit{decisions} are suboptimal for a significant percentage (28\%) of feature values. 

As shown in Figure \ref{fig:casestudyCART4}, when CART is permitted to utilize more splits up to a tree depth of 4, it is able to nearly recover the optimal decision boundary. Even though each individual split taken by CART has less value for decision-making, the splits in combination finely partition the feature space into small enough regions that the predicted cost vectors are highly accurate within each region. Therefore, when trained to a significant depth, CARTs -- and more generally, decision trees -- potentially have a high enough model complexity to achieve near perfect predictions which translate into near perfect decisions. However, in settings with limited training data, it is no longer possible to train decision trees to a suitably high depth, as a sufficient number of training observations per leaf are required to estimate the leaf cost predictions accurately. Therefore, in these settings, maximizing the contribution of each decision tree split to optimal decision-making becomes a higher priority. Moreover, lower depth decision trees are often preferred for their interpretability and reduced risk of overfitting.

Figure \ref{fig:casestudyCARTerrors} assesses the decisions from the SPOT and CART algorithms when trained to different tree depths. The decisions are scored on a held out set of data using the metric of ``normalized extra travel time'', defined as the cumulative SPO loss normalized by the cumulative optimal decision costs. $\sum_{i=1}^{n}\ell_{SPO}(\hat{c}_i,c_i)/\sum_{i=1}^{n}z^*(c_i)$.
Unsurprisingly, the SPO Tree achieves zero decision error at all training depths since it correctly identified the decision boundary at depth 1. By comparison, the CART algorithm exhibits comparatively high decision error at depths 1-3 and only begins to reach a decision error near zero at depth 4. Therefore, the SPO Tree achieves high quality decisions while also being significantly less complex than the CART tree required for comparable decision quality. We show in Section \ref{sec:Results} that this behavior is consistently observed across a range of synthetic and real datasets.




\pdfoutput=1
\section{Methodology} \label{sec:spotmethods}
 We now propose several algorithms for training decision trees using the SPO loss function, and we call the resulting models SPO Trees (SPOTs). The objective of any decision tree training algorithm is to partition the training observations into $L$ leaves, $R_1,...,R_{L} := R_{1:L}$, whose predictions collectively minimize a given loss function:
\begin{align}
\min_{R_{1:L} \in \mathcal{T}} \frac{1}{n}  \sum_{l=1}^{L} \left(\min_{\hat{c_l}} \sum_{i \in R_l } \ell(\hat{c_l},  c_i)  \right) \label{eqn:spoopt}
\end{align}
Above, the constraint $R_{1:L} \in \mathcal{T}$ indicates that the allocation of observations to leaves must follow the structure of a decision tree (i.e., determined through repeated splits on the feature components). The CART algorithm greedily selects tree splits which individually minimize this objective with respect to mean squared error prediction loss \citep{breiman1984classification}. More recently, integer programming strategies have been proposed for optimally solving \eqref{eqn:spoopt} with respect to classification loss \citep{bertsimas2017optimal,gunluk2018optimal,verwer2019learning,hu2019optimal,aghaei2020learning}. We next describe tractable extensions of these greedy and integer programming methodologies from the literature to train decision trees using SPO loss, which has been shown to have favorable generalization bounds in several settings \citep{el2019generalization}.

\citet{elmachtoub2017smart} note that training machine learning models under SPO loss is likely infeasible due to the loss function being nonconvex and discontinuous in the predicted cost vectors. However, we show that optimization problem (\ref{eqn:spoopt}) for training decision trees under SPO loss can be greatly simplified through Theorem \ref{min_within_partition_SPO}, which states that the average of the cost vectors corresponding to a leaf node minimizes the SPO loss in that leaf node. 
\begin{theorem}\label{min_within_partition_SPO}
Let $\bar{c}_l := \frac{1}{|R_l|} \sum_{i\in R_l} c_i$ denote the average cost of all observations within leaf $l$. If  $\bar{c}_l$ has a unique minimizer in its corresponding decision problem, then $\bar{c}_l$ minimizes within-leaf SPO loss. More simply, if $|W^*(\bar{c}_l)|=1$, then $\bar{c}_l = \argmin_{\hat{c_l}}\sum_{i \in R_l } \ell_{SPO}(\hat{c_l}, c_i)$.
\end{theorem}

\proof{Proof:}
Let $\bar{c}_l$ be defined as stated in the theorem. We will show that the within-leaf SPO loss associated with predicting $\bar{c}_l$ lower bounds that of predicting any other feasible cost vector $\hat{c}_l \in \mathbb{R}^d$. Let $N_l = |R_l|$ denote the number of observations within leaf $l$. The following holds for any $\hat{c}_l \in \mathbb{R}^d$:
\begin{align*}
& \frac{1}{N_l}\sum_{i \in R_l } \ell_{SPO}(\bar{c}_l,  c_i) - \frac{1}{N_l}\sum_{i \in R_l } \ell_{SPO}(\hat{c}_l,  c_i)  \\
  ~=~ & \frac{1}{N_l}\sum_{i \in R_l }  \max_{w \in W^*(\bar{c}_l)} \{ c_i^Tw\} - \frac{1}{N_l}\sum_{i \in R_l } \max_{w \in W^*(\hat{c}_l)} \{ c_i^Tw\} \\
  ~=~ & \frac{1}{N_l}\sum_{i \in R_l } c_i^T  w^*(\bar{c}_l) - \frac{1}{N_l}\sum_{i \in R_l } \max_{w \in W^*(\hat{c}_l)} \{c_i^T w\} \quad \left(W^*(\bar{c}_l) = \{ w^*(\bar{c}_l) \} \text{ is a singleton} \right) \\
  ~\leq~ & \frac{1}{N_l}\sum_{i \in R_l } c_i^T  w^*(\bar{c}_l) - \max_{w \in W^*(\hat{c}_l)} \bigg\{ \frac{1}{N_l}\sum_{i \in R_l } c_i^T w\bigg\} \\
  ~=~ & \bar{c}_l^T  w^*(\bar{c}_l) -  \max_{w \in W^*(\hat{c}_l)} \{ \bar{c}_l^T w\}  \\
  ~\leq~ & 0 \quad \left(\text{by definition of } w^*(\bar{c}_l) \right)
\end{align*} 

We have thus demonstrated that $\bar{c}_l$ achieves a within-leaf SPO loss lower or equal to that of any other cost vector $\hat{c}_l \in \mathbb{R}^d$, thereby proving the theorem. 
\halmos


Note that the optimal solution to the underlying decision problem has a unique solution except in a few degenerate cases (e.g., the supplied cost vector is the zero vector). To ensure that these degenerate cases have measure 0, it is sufficient to assume that the marginal distribution of $c$ given $x$ is continuous and positive on $\mathbb{R}^{d}$. Empirically, to guarantee uniqueness of an optimal solution, one can simply add a small noise term to every cost vector in the training set. Therefore, in what follows, we assume that $W^*(\bar{c}_l)$ is a singleton for any feasible $\bar{c}_l$ and utilize Theorem \ref{min_within_partition_SPO} throughout. 

Theorem \ref{min_within_partition_SPO}  expresses that the cost vector which minimizes within-leaf SPO loss may be expressed in closed form as the average of the cost vectors belonging to the given leaf. We utilize this information to greatly simply optimization problem (\ref{eqn:spoopt}):
\begin{align}
\begin{aligned}[b]
\label{eqn:spoopt2}
 & \min_{R_{1:L} \in \mathcal{T}} \frac{1}{n}  \sum_{l=1}^{L} \left(\min_{\hat{c_l}} \sum_{i \in R_l } \ell_{SPO}(\hat{c_l},  c_i)  \right) \\
=& \min_{R_{1:L} \in \mathcal{T}} \frac{1}{n}  \sum_{l=1}^{L} \sum_{i \in R_l } \left( c_i^T w^*(\bar{c}_l)  - z^*(c_i)  \right)\,.
\end{aligned}
\end{align}

\subsection{SPOT: Recursive Partitioning Approach} \label{sec:recpar}

To obtain a quick and reliable solution to optimization problem (\ref{eqn:spoopt2}), we propose using \textit{recursive partitioning} to train SPO Trees with respect to the above objective function.  CART employs the same procedure to find decision trees which approximately minimize training set prediction error. Define $x_{i,j}$ as the $j$-th feature component corresponding to the $i$-th training set observation. Beginning with the entire training set, consider a decision tree split $(j,s)$ represented by a splitting feature component $j$ and split point $s$ which partitions the observations into two leaves: 
 \begin{equation*}
 \textstyle
 R_1(j,s) = \{i \in [n] \mid x_{i,j} \leq s \} \,\,\, \text{and} \,\,\, R_2(j,s) = \{i \in [n]  \mid  x_{i,j} > s \} \,,
 \end{equation*}
if variable $j$ is numeric, or 
 \begin{equation*}
 R_1(j,s) = \{i \in [n]  \mid  x_{i,j} = s \} \,\,\, \text{and} \,\,\, R_2(j,s) = \{i \in [n]  \mid  x_{i,j} \neq s \} \,,
 \end{equation*}
if variable $j$ is categorical. Here, we define $[n]$ as shorthand notation for the set $\{1,2,...,n\}$. The first split of the decision tree is chosen by computing the pair $(j,s)$ which minimize the following optimization problem:
\begin{align}
\label{eqn:greedyspo}
\min\limits_{j,s}& \frac{1}{n} \left( \sum_{i \in R_1(j,s)} \left( c_i^T w^*(\bar{c}_l)  - z^*(c_i)  \right)  +  \sum_{i \in R_2(j,s)} \left( c_i^T w^*(\bar{c}_l)  - z^*(c_i)  \right) \right)\,.
\end{align}In words, the training procedure ``greedily'' selects the single split whose resulting decisions obtain the best SPO loss on the training set. Problem (\ref{eqn:greedyspo}) can be solved by computing the objective function value associated with every feasible split $(j,s)$ and selecting the split with the lowest objective value. Leveraging Theorem \ref{min_within_partition_SPO}, a split's objective value may be determined by (1) partitioning the training observations according to the split, (2) determining the average cost vectors $\bar{c}_1$ and $\bar{c}_2$ and associated decisions $w^*(\bar{c}_1)$ and $w^*(\bar{c}_2)$ in each leaf, (3) computing the SPO loss in each leaf resulting from the decisions, and (4) adding the SPO losses together and dividing by $n$. We observe empirically that the computation of a split's objective value is very fast due to the decision oracle $w^*(\cdot)$ only needing to be called once in each partition. Checking all possible split points $s$ associated with \textit{continuous} feature components $j$ may be computationally prohibitive, so instead we recommend the following heuristic. All unique values of the continuous feature observed in the training data are sorted, and the consideration set of potential split points is determined through only considering certain quantiles of the feature values.

After a first split is chosen, the greedy split selection approach is then recursively applied in the resulting leaves until one of potentially several stopping criteria is met. Common stopping criteria to be specified by the practitioner include a maximum depth size for the tree and/or a minimum number of training observations per leaf. The decision tree pruning procedure from \citet{breiman1984classification} (using SPO loss as the pruning metric) may be further applied to reduce model complexity and prevent overfitting.

\subsection{SPOT: Integer Programming Approach} \label{sec:ipapproach}
We also consider using integer programming to solve optimization problem (\ref{eqn:spoopt}) to optimality for training decision trees using SPO loss. Here we leverage the simplified form (\ref{eqn:spoopt2}) of optimization problem (\ref{eqn:spoopt}) derived using Theorem \ref{min_within_partition_SPO}. We show that the optimization problem (\ref{eqn:spoopt2}) may be equivalently expressed as a mixed integer linear program (MILP). MILPs are generally regarded as being computationally feasible in many settings due to an incredible increase in the computational power and sophistication of mixed-integer optimization solvers such as Gurobi and CPLEX over the past decade. Let $r_{il}$ denote a binary variable which indicates whether training observation $i$ belongs to leaf $R_l$. Then, 
\begin{align*}
\min_{R_{1:L} \in \mathcal{T}} \frac{1}{n}  \sum_{l=1}^{L} \sum_{i \in R_l } \left( c_i^T w^*(\bar{c}_l)  - z^*(c_i)  \right)
=  \min_{r_{1:L} \in \mathcal{T}} \frac{1}{n}  \sum_{l=1}^{L} \sum_{i=1}^{n} r_{il} \left( c_i^T w^*(\bar{c}_l)  - z^*(c_i)  \right) \,. 
\end{align*}
Recall that the constraint $r_{1:L} \in \mathcal{T}$ indicates that the allocation of observations to leaf nodes must follow the structure of a decision tree (i.e., determined through repeated splits on the feature components). There have been several frameworks proposed in the literature for encoding decision trees using integer and linear constraints \citep{bertsimas2017optimal, gunluk2018optimal, verwer2019learning,aghaei2020learning}. We have chosen to apply the framework proposed by \citet{bertsimas2017optimal}, as it naturally accommodates both continuous and categorical splits and also automatically pools together leaf nodes which do not contribute to minimizing the objective function (provided a small regularization parameter is introduced). 
We provide the complete formulation of $r_{1:L} \in \mathcal{T}$ as integer and linear constraints in Appendix \ref{sec:milpdet}. 

Define $M_1 := \max\{\max_{i,w \in S} c_i^T w, 0\}$ and $M_2 := \max\{\max_{i,w \in S} -c_i^T w, 0\}$ as sufficiently large nonnegative constants. We assume that the decision feasibility constraint set $S$ is bounded, guaranteeing that $M_1$ and $M_2$ are finite. Note that $M_1$ and $M_2$ may also be defined in terms of $z^*(\cdot)$ as $\max\{\max_{i} -z^*(-c_i), 0\}$ and $\max\{\max_{i} -z^*(c_i), 0\}$, respectively. Theorem \ref{SPOIP} shows that optimization problem  (\ref{eqn:spoopt2}) may be equivalently expressed as a mixed integer linear program (MILP) and therefore can be tractably solved to optimality for a modest number of integer variables. 

\begin{theorem}\label{SPOIP}
Assume that the decision feasibility constraints $w \in S$ consist of only linear and integer constraints and that $S$ is bounded. Then, optimization problem (\ref{eqn:spoopt2}) may be equivalently expressed as the following MILP:
\begin{mini}[2] 
    {r,w,y}{ \frac{1}{n}   \sum_{l=1}^{L} \sum_{i=1}^{n} y_{il}  - \sum_{i=1}^{n} z^*(c_i) }
    {}{}
    \addConstraint{y_{il}}{\geq c_i^T  w_l - M_1(1-r_{il}),\quad}{\forall i\in \{1...n \}, l\in \{1...L \}}
    \addConstraint{y_{il}}{\geq -M_2 r_{il}}{\forall i\in \{1...n \}, l\in \{1...L \}}
    \addConstraint{w_l}{\in S}{\forall l\in \{1...L \}}
	\addConstraint{r_{il}}{\in \mathcal{T}}{\forall i\in \{1...n \}, l\in \{1...L \}}
\end{mini} \label{eqn:themilp}
\end{theorem}
\proof{Proof:}
 Let $N_l = |R_l|$ denote the number of observations within leaf $l$. We first perform the following algebraic operations starting with optimization problem (\ref{eqn:spoopt2}):
\begin{align*}
	&  \min_{R_{1:L} \in \mathcal{T}} \frac{1}{n}  \sum_{l=1}^{L} \sum_{i \in R_l } \left( c_i^T w^*(\bar{c}_l)  - z^*(c_i)  \right) \\ 
    =&  \min_{R_{1:L} \in \mathcal{T}} \frac{1}{n}   \sum_{l=1}^{L} \left( N_l \bar{c}_l^T w^*(\bar{c}_l) - \sum_{i \in R_l } z^*(c_i) \right) \\
    =&  \min_{R_{1:L} \in \mathcal{T}} \frac{1}{n}   \sum_{l=1}^{L} \left( N_l \min_{w_l \in S} \{\bar{c}_l^T w_l\} - \sum_{i \in R_l } z^*(c_i) \right) \\
    =&  \min_{\substack{R_{1:L} \in \mathcal{T} \\ w_{1:L} \in S}}\frac{1}{n}   \sum_{l=1}^{L} \left( N_l \bar{c}_l^T w_l - \sum_{i \in R_l } z^*(c_i) \right) \\
    =&  \min_{\substack{R_{1:L} \in \mathcal{T} \\ w_{1:L} \in S}}\frac{1}{n}   \sum_{l=1}^{L} \sum_{i \in R_l } \left( c_i^T w_l - z^*(c_i) \right)\,.
\end{align*}
Let $r_{il}$ denote a binary variable which indicates whether training observation $i$ belongs to leaf $R_l$. Then, 
\begin{align*}
     &\min_{\substack{R_{1:L} \in \mathcal{T} \\ w_{1:L} \in S}}\frac{1}{n}   \sum_{l=1}^{L} \sum_{i \in R_l } \left( c_i^T w_l - z^*(c_i) \right)\\ ~=~ & \min_{\substack{r_{1:L} \in \mathcal{T} \\ w_{1:L} \in S}}\frac{1}{n}   \sum_{l=1}^{L} \left( \sum_{i=1}^{n}  r_{il} c_i^T w_l \right) - \sum_{i=1}^{n} z^*(c_i)\\
     ~=~ & \min_{\substack{r_{1:L} \in \mathcal{T} \\ w_{1:L} \in S \\ y_{1:L}}}\frac{1}{n}   \sum_{l=1}^{L} \sum_{i=1}^{n} y_{il}  - \sum_{i=1}^{n} z^*(c_i) \,,
\end{align*}
where in the last step we add the constraint that $y_{il} = r_{il} c_i^T  w_l$ for every $i$ and $l$. First, note that this constraint may be equivalently expressed as $y_{il} \geq r_{il} c_i^T  w_l$, as $y_{il}$ will always be set equal to its minimum feasible value ($r_{il} c_i^T  w_l$) since it is being minimized in the objective function. However, this constraint is still not linear since it involves the multiplication of two decision variables $r_{il}$ and $w_l$. We may rewrite it as the two linear constraints below:
\begin{equation*}
    y_{il} \geq c_i^T  w_l - M_1(1-r_{il}) ~~~\text{and}~~~ y_{il} \geq -M_2r_{il} \,.
\end{equation*}
Above, $M_1$ and $M_2$ are constants which upper bound $c_i^T  w_l$ and $-c_i^T  w_l$, respectively, for all $i\in \{1,2,\dots,n\}$ and $w_l \in S$. We therefore define $M_1 := \max\{\max_{i,w \in S} c_i^T w, 0\}$ and $M_2 := \max\{\max_{i,w \in S} -c_i^T w, 0\}$ which are finite due to $S$ being bounded. 
Note that when the cost vectors are all nonnegative (nonpositive), then $M_2 = 0$ ($M_1 = 0)$ assuming the decision variables $w$ are nonnegative for all feasible $w \in S$. Thus, the optimization problem for training decision trees under SPO loss may be written as the following mixed integer linear program:
\begin{mini*}[2] 
    {r,w,y}{ \frac{1}{n}   \sum_{l=1}^{L} \sum_{i=1}^{n} y_{il}  - \sum_{i=1}^{n} z^*(c_i) }
    {}{}
    \addConstraint{y_{il}}{\geq c_i^T  w_l - M_1(1-r_{il}),\quad}{\forall i\in \{1...n \}, l\in \{1...L \}}
    \addConstraint{y_{il}}{\geq -M_2 r_{il}}{\forall i\in \{1...n \}, l\in \{1...L \}}
    \addConstraint{w_l}{\in S}{\forall l\in \{1...L \}}
	\addConstraint{r_{il}}{\in \mathcal{T}}{\forall i\in \{1...n \}, l\in \{1...L \}}
\end{mini*}
\halmos

Empirically, we have noticed a significant computational speed up in solving the MILP if it is warm started with the solution recovered from the greedy algorithm. Furthermore, since the greedy algorithm produces a \textit{feasible} solution for the MILP, then the MILP is guaranteed to recover a solution which is at least as optimal as the greedy solution, even if the MILP solver is prematurely terminated. Therefore, in settings where training the MILP to optimality is computationally infeasible, we recommend warm-starting the MILP algorithm with the greedy algorithm and using the MILP as a ``solution improvement tool'', allowing the solver to continually improve the solution until being terminated after it has exceeded a specified time limit. This is the procedure we employ in our numerical experiments, specifying a maximum time limit of 12 hours. Other strategies we employ for improving the computation time of the SPOT MILP approach as well as other implementation details (including regularization procedures to prevent overfitting) may be found in Appendix \ref{sec:milpdet2}.

\subsection{SPO Forests}

We also consider training an \textit{ensemble} of SPO Trees, a methodology which we call \textit{SPO Forests}. SPO Forests are constructed using (greedy) SPO Trees through the same procedure as random forests are constructed using CARTs. Random forests are known to have less variance than individual decision trees, at the price of sacrificing interpretability \citep{friedman2001elements}. To construct an SPO Forest, $B$ SPO Trees are trained on bootstrapped samples of the training dataset, where $B$ represents the number of desired trees in the SPO Forest. To further reduce the correlation between trees, we implement feature bagging, defined as only considering a random subset of features when deciding splits in the learning process. When presented with a new feature vector $x_{new}$, the cost vectors predicted by the SPO Trees are averaged, and the SPO Forest returns the optimal decision associated with this average cost vector.

\pdfoutput=1
\section{Experimental Results} \label{sec:Results}
\subsection{Noisy Shortest Path:}
We first study the empirical performance of SPO Trees and SPO Forests on a synthetic dataset for the shortest path problem studied in \citet{elmachtoub2017smart}. 
For sake of comparison, we also train CART decision trees and CART random forests on the same datasets using the loss function of mean squared prediction error. The shortest path problem considered is with respect to a 4 x 4 grid network consisting of edges (``roads'') which are only directed north and east. The driver starts at the southwest corner of the grid, and the goal of the driver is to travel to the northeast corner via the shortest path available. The costs (``travel times'') associated with the 24 edges of the network are unknown but can be predicted using five numerical features. Datasets of $n \in \{200, 10000\}$ feature-cost pairs are generated by (1) sampling $n$ feature vectors $x_1, ..., x_n$ each from a $Uniform(0,1)^p$ distribution where $p = 5$, (2)  sampling matrix $B\in \{0,1\}^{d \times p}$ by sampling each entry $B_{k,j}$ from $Bernoulli(1,0.5)$, and (3) computing each feature vector $x_i$'s associated cost vector $c_i$ according to $c_{ik} = \left( \frac{1}{\sqrt{p}} \left(  Bx_i\right)_k +1 \right)^{deg}\cdot \varepsilon_i^k$, where $(Bx_i)_k$ denotes the $k$th component of $Bx_i$, $deg$ is a fixed positive integer that controls the amount of nonlinearity present in the mapping from features to cost vectors, 
and $\varepsilon_i^k$ are multiplicative i.i.d. noise terms sampled from $Uniform([1- \bar{\varepsilon}, 1 + \bar{\varepsilon}])$  for some parameter $\bar{\varepsilon} \geq 0$. 
We consider several combinations of the parameters $n$, $deg$ and $\bar{\varepsilon}$. For each combination of parameters, 10 datasets are generated with uniquely sampled $B$ matrices. The algorithms are tested on a set of 1000 observations generated using the same $B$ as the training set. Algorithmic performance on the test set is assessed with respect to normalized extra travel time defined in Section (\ref{sec:illustrative}), which is equivalent to (normalized) SPO loss. 

All trees and forests are trained using a minimum leaf size of 20 observations. To prevent overfitting, SPOTs and CART trees are pruned on a validation set consisting of 20\% of the training data using the pruning algorithm from \citet{breiman1984classification}. The forest algorithms are trained using $B = 100$ trees with no depth limit, and the number of features $f \in \{2,3,4,5\}$ to use in feature bagging is tuned using the validation set above. 

\begin{figure}
	\centering
	\begin{subfigure}{0.47\linewidth}
		\centering
		\includegraphics[width=0.97\linewidth]{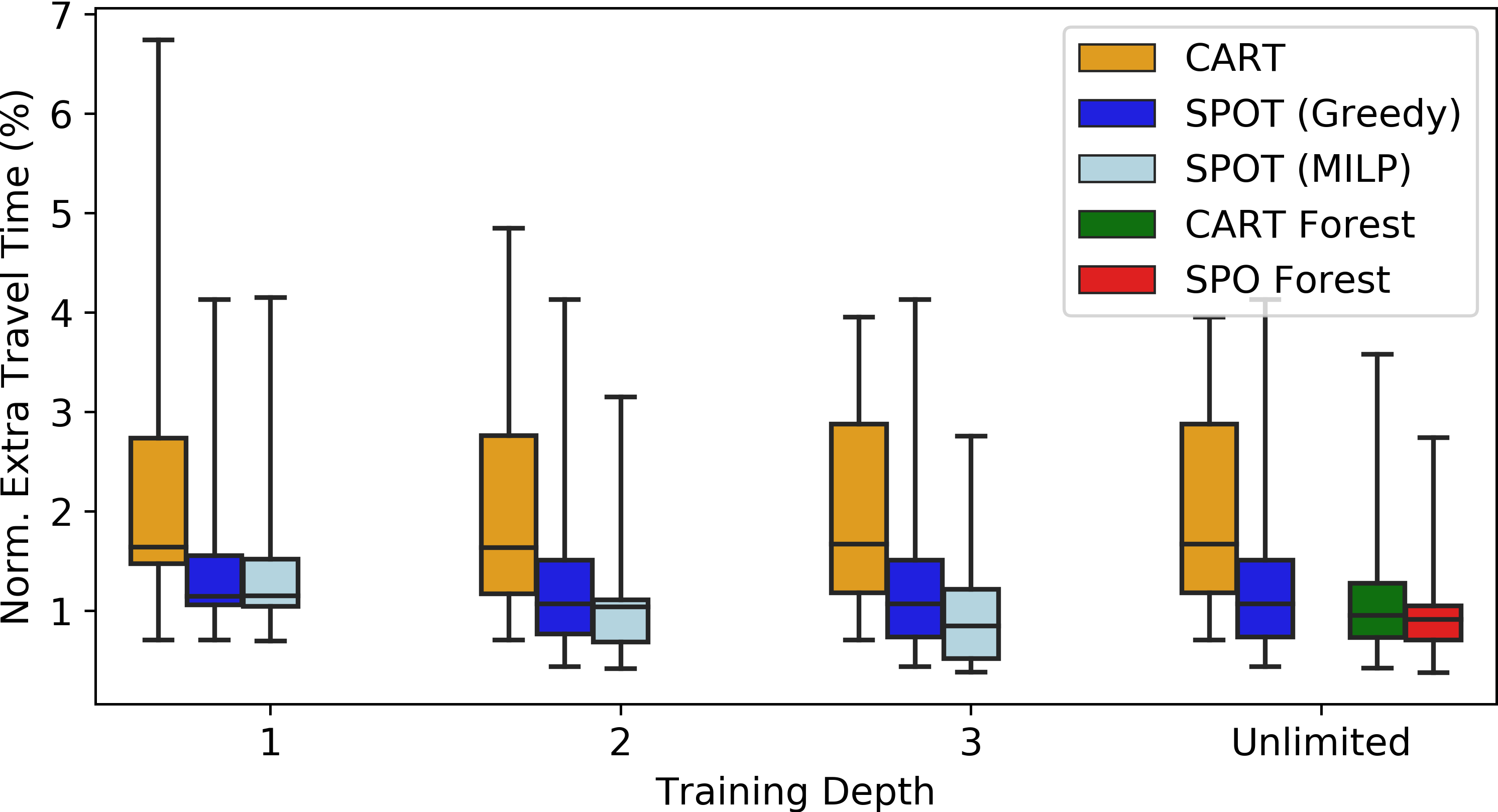}
		\caption{\small $deg = 2$, $\bar{\varepsilon} = 0$}
		\vspace{0.1cm}
		\label{fig:tr200_eps0_deg2}
	\end{subfigure}%
	\begin{subfigure}{0.47\linewidth}
		\centering
		\includegraphics[width=0.97\linewidth]{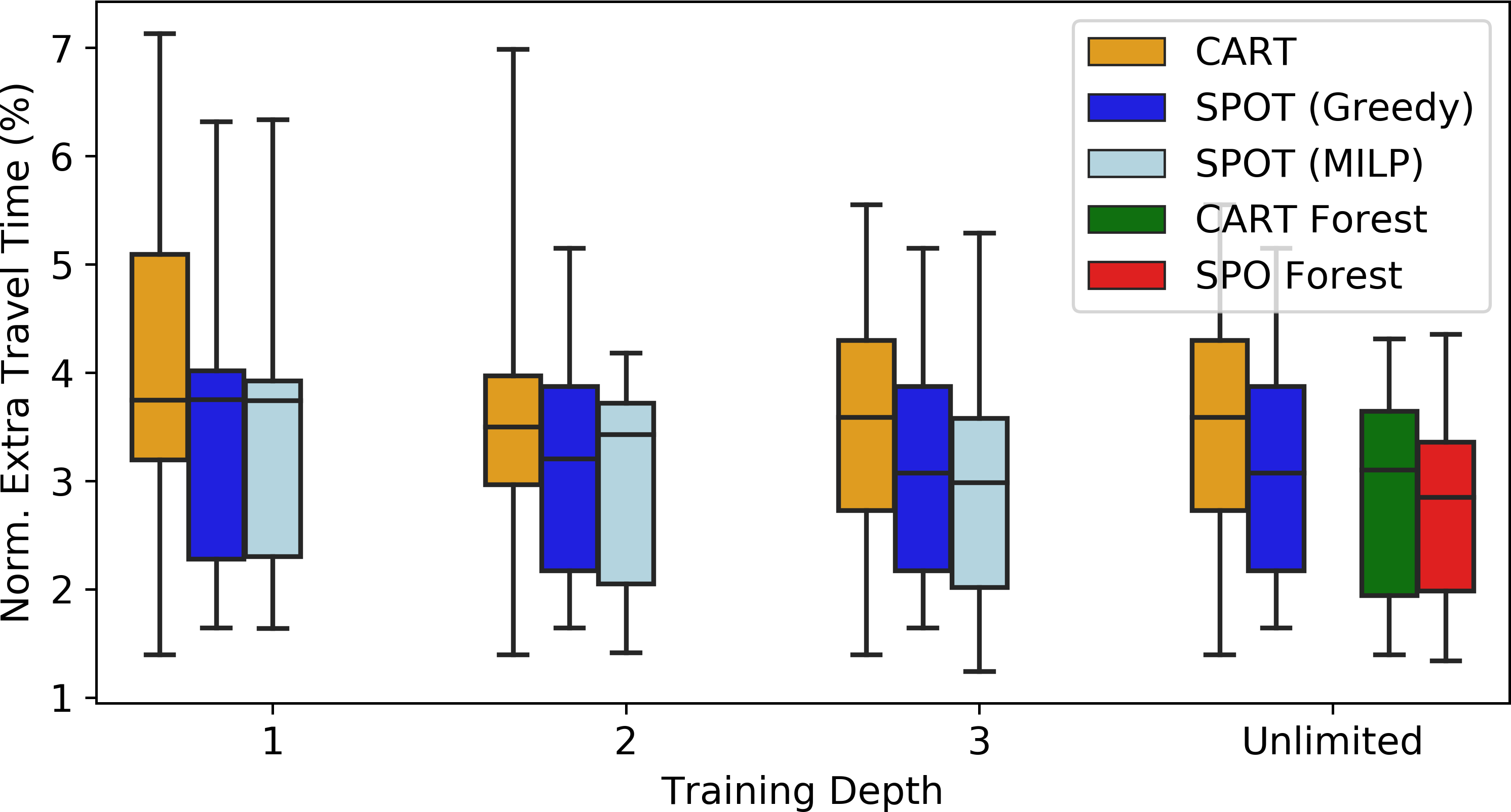}
		\caption{\small $deg = 2$, $\bar{\varepsilon} = 0.25$}
		\vspace{0.1cm}
		\label{fig:tr200_eps0.25_deg2}
	\end{subfigure}
	\\
	\begin{subfigure}{0.47\linewidth}
		\centering
		\includegraphics[width=0.97\linewidth]{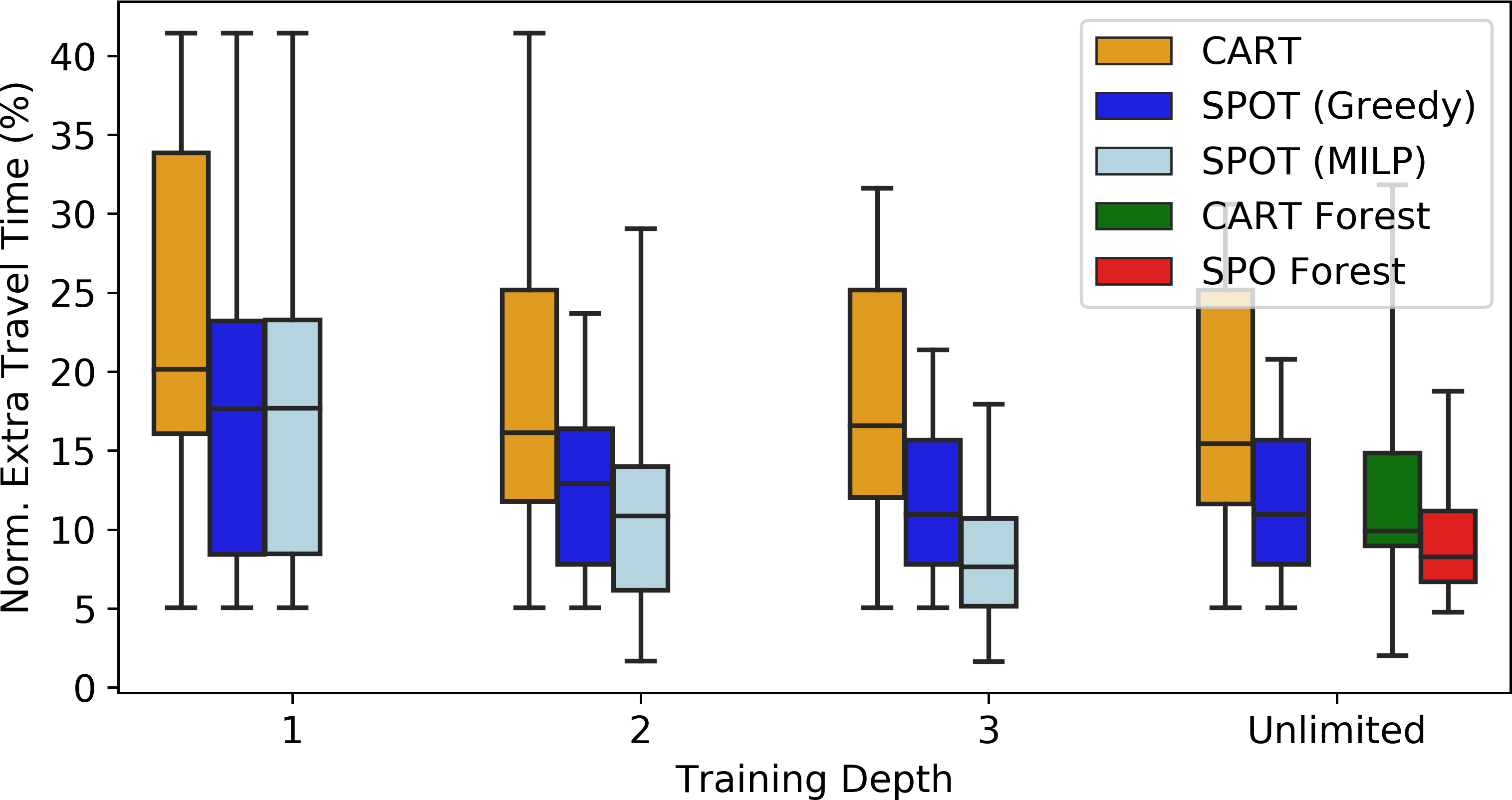}
		\caption{\small $deg = 10$, $\bar{\varepsilon} = 0$}
		\vspace{0.1cm}
		\label{fig:tr200_eps0_deg10}
	\end{subfigure}%
	\begin{subfigure}{0.47\linewidth}
		\centering
		\includegraphics[width=0.97\linewidth]{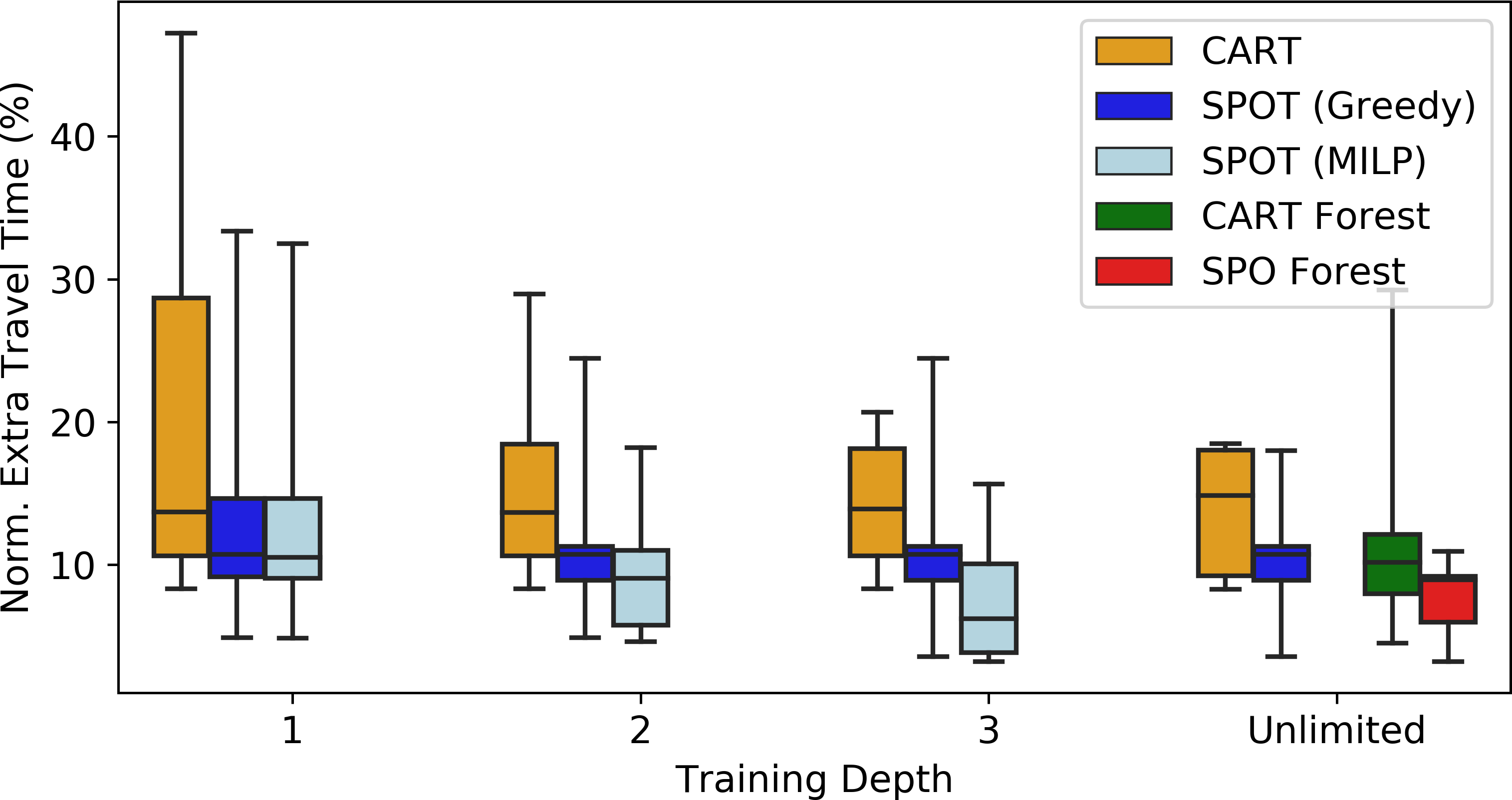}
		\caption{\small $deg = 10$, $\bar{\varepsilon} = 0.25$}
		\vspace{0.1cm}
		\label{fig:tr200_eps0.25_deg10}
	\end{subfigure}
	\caption{Test set normalized extra travel times on 10 different shortest path datasets of size $n = 200$.}
	\label{fig:tr200}
\end{figure}

We begin by considering the performance of the decision tree algorithms in an experimental setting with limited training data. We fix the number of training observations at $n = 200$ and vary the experimental parameters $deg \in \{2,10\}$ and $\bar{\varepsilon} \in \{0, 0.25\}$.  We evaluate the performance of SPOT and CART trees when trained to fixed depths of 1, 2, and 3 on the training set. We also include the performance of the SPOT and CART algorithms when imposing \textit{no} restrictions on their training depth (but still employing the pruning algorithm to prevent overfitting). Note that the SPOT MILP approach requires a fixed training depth and is therefore not included in the algorithms with no depth restriction. Figure \ref{fig:tr200} visualizes the test-set performance of the SPOT algorithms and benchmarks on the shortest path problem with $n = 200$ observations for all combinations of experimental parameters $\deg$ and $\bar{\epsilon}$.

We observe that SPO Trees significantly outperform CART in all settings of the experimental parameters. In particular, the greedy SPOT algorithm achieves percentage improvements in normalized extra travel time over the CART algorithm of 26.7\%, 26.8\%, 23.1\%, and 23.6\% when both are trained to depths of 1, 2, 3, and unrestricted depth, respectively (with the above percentage improvements averaged across the four combinations of $deg$ and $\bar{\epsilon}$). In general, the SPO Trees trained to depth 1 often achieve a lower SPO loss than the CART trees trained with \textit{unrestricted} depth. Therefore, the SPO Trees lead to better decisions than CART while also being more concise and therefore more interpretable. The failure of CART to achieve competitive decision performance can be explained by its focus on prediction (rather than decision) error coupled with the limited amount of training data. Recall that a minimum of 20 training observations are required to be mapped to each leaf of the decision trees -- this constraint is imposed to ensure that the costs within each leaf are estimated with sufficient accuracy. Even with no depth limit, we observe empirically that the CART trees cannot be trained past a depth of 4 without the minimum leaf size criterion being satisfied. Therefore, in small data settings, the number of splits which decision trees may utilize are limited, and thus it becomes imperative to maximize the contribution of each split towards decision quality. A comparison of the random forest algorithms mirrors these findings -- forests of SPO Trees consistently outperform forests of CART trees by 20.5\% averaged across the four parameter settings, notably also achieving less variance in performance (i.e., boxplot width) than CART trees. The SPO Tree MILP approach offers additional improvements in decision quality when compared to the SPOT greedy approach, outperforming even the random forest algorithms in some cases.  

\begin{figure}
	\centering
	\begin{subfigure}{0.47\linewidth}
		\centering
		\includegraphics[width=0.97\linewidth]{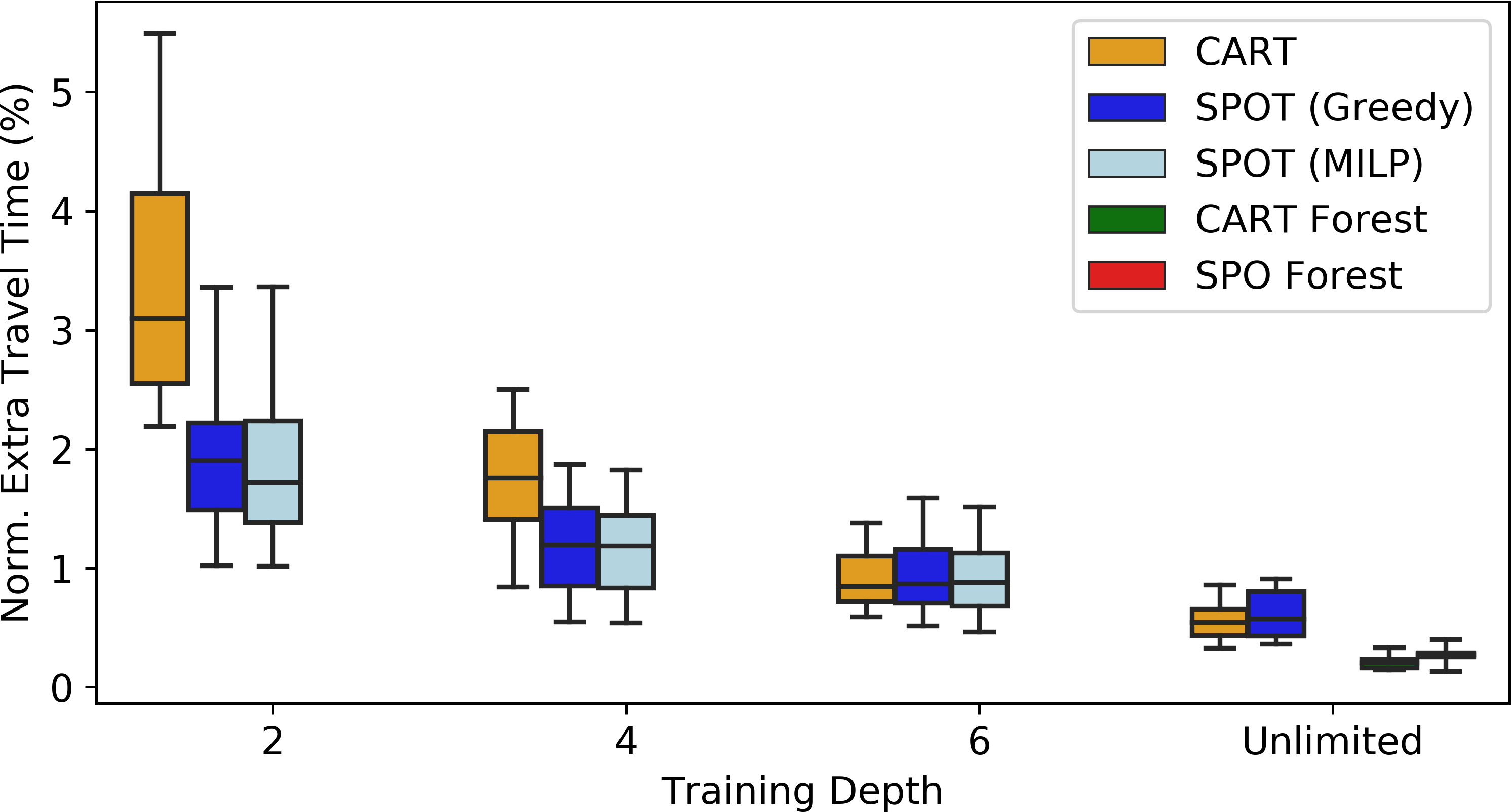}
		\caption{\small $deg = 2$, $\bar{\varepsilon} = 0$}
		\vspace{0.1cm}
		\label{fig:tr10000_eps0_deg2}
	\end{subfigure}%
	\begin{subfigure}{0.47\linewidth}
		\centering
		\includegraphics[width=0.97\linewidth]{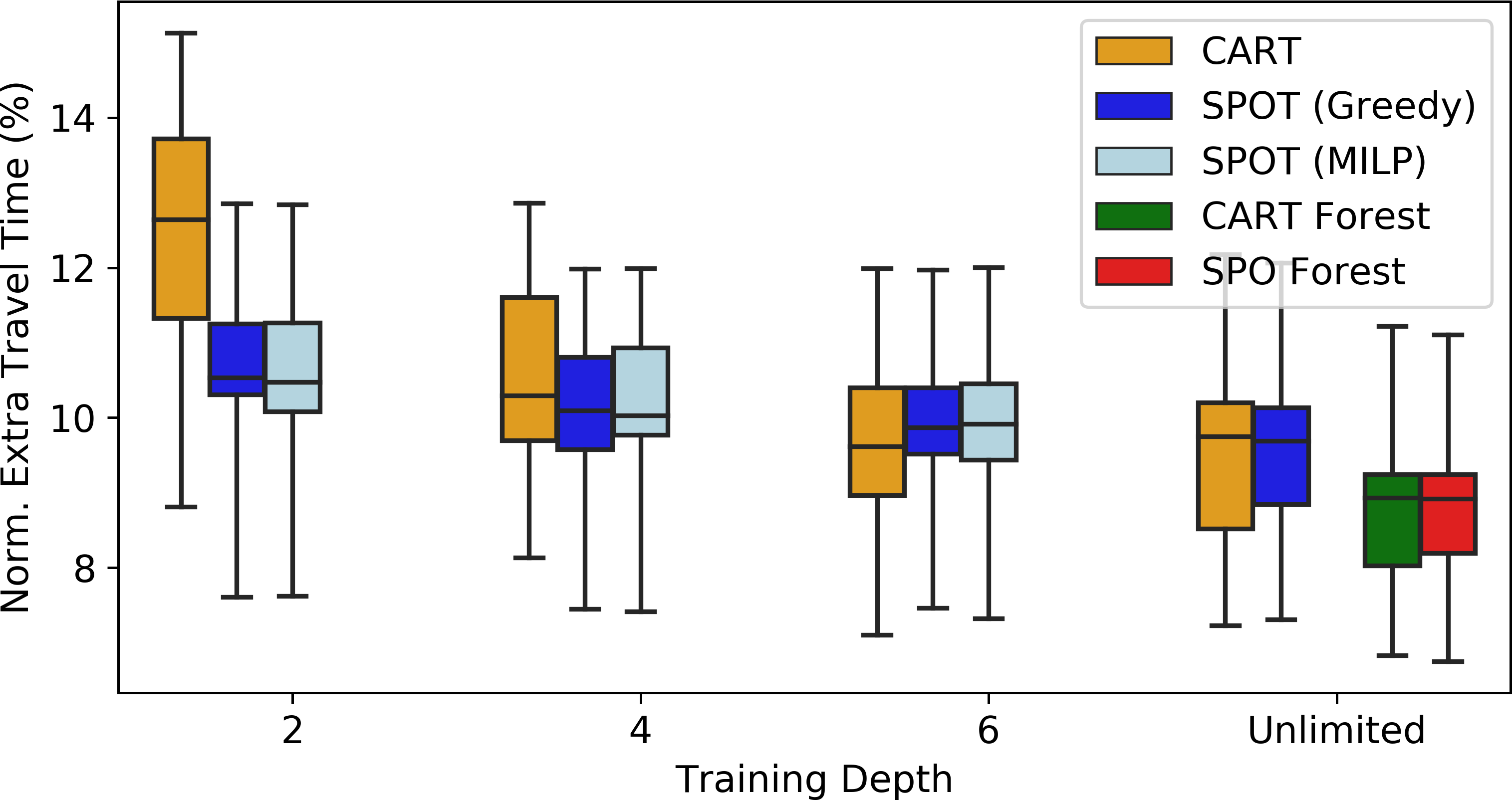}
		\caption{\small $deg = 2$, $\bar{\varepsilon} = 0.5$}
		\vspace{0.1cm}
		\label{fig:tr10000_eps0.5_deg2}
	\end{subfigure}
	\\
	\begin{subfigure}{0.47\linewidth}
		\centering
		\includegraphics[width=0.97\linewidth]{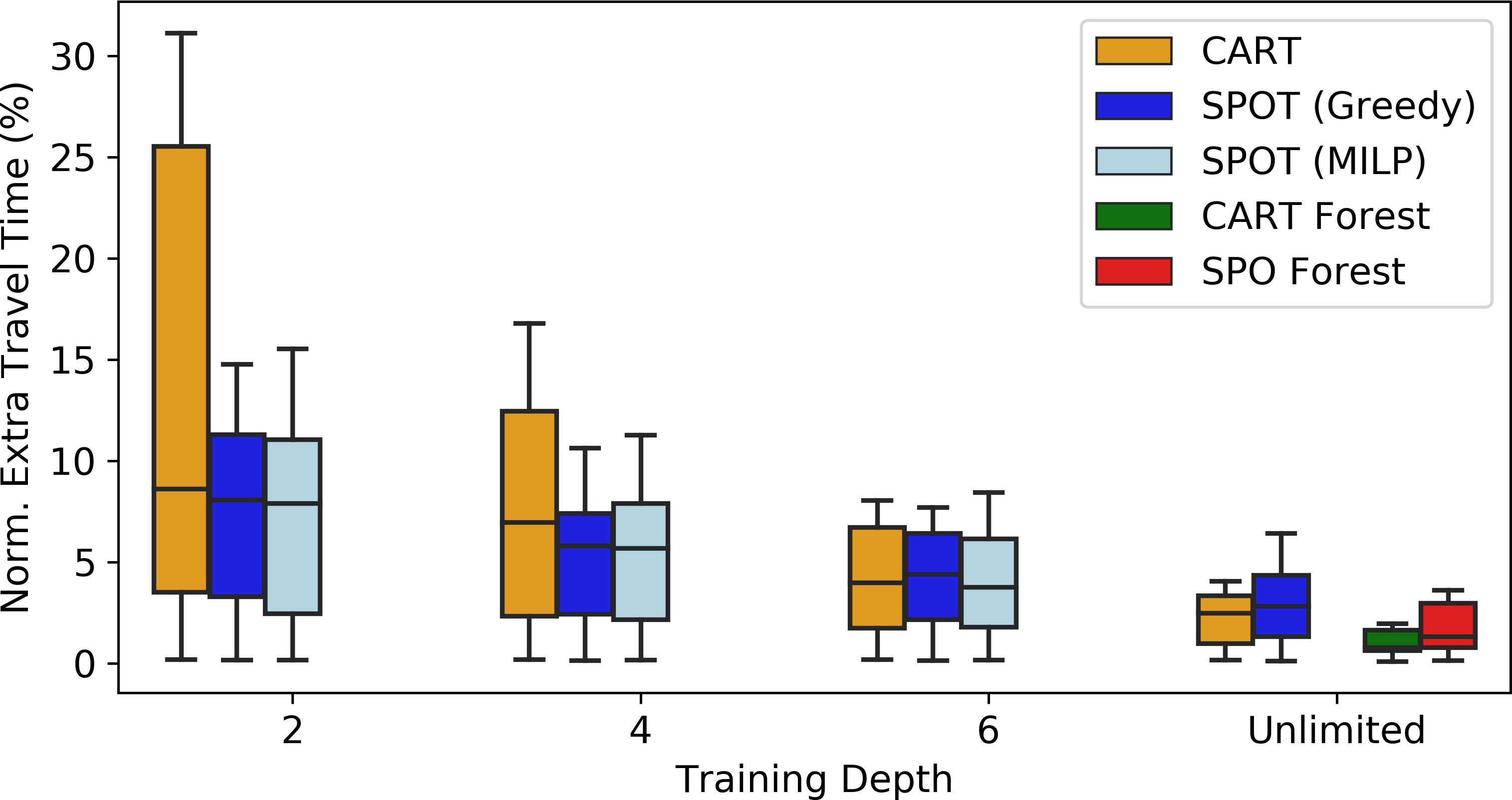}
		\caption{\small $deg = 10$, $\bar{\varepsilon} = 0$}
		\vspace{0.1cm}
		\label{fig:tr10000_eps0_deg10}
	\end{subfigure}%
	\begin{subfigure}{0.47\linewidth}
		\centering
		\includegraphics[width=0.97\linewidth]{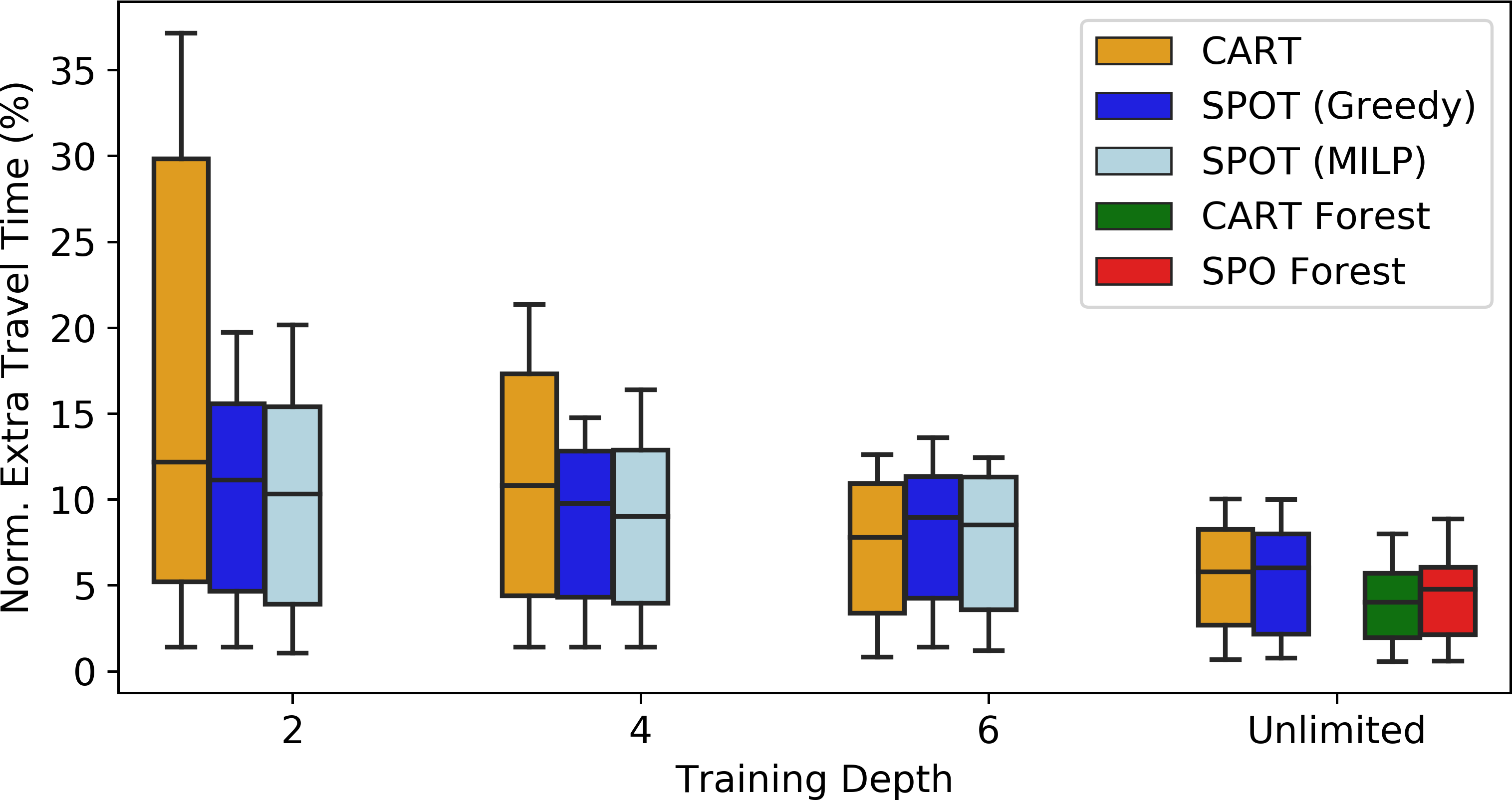}
		\caption{\small $deg = 10$, $\bar{\varepsilon} = 0.5$}
		\vspace{0.1cm}
		\label{fig:tr10000_eps0.5_deg10}
	\end{subfigure}
	\caption{Test set normalized extra travel times on 10 different shortest path datasets of size $n = 10000$.}
	\label{fig:tr10000}
\end{figure}

\begin{figure}
	\centering
	\begin{subfigure}{0.47\linewidth}
		\centering
		\includegraphics[width=0.97\linewidth]{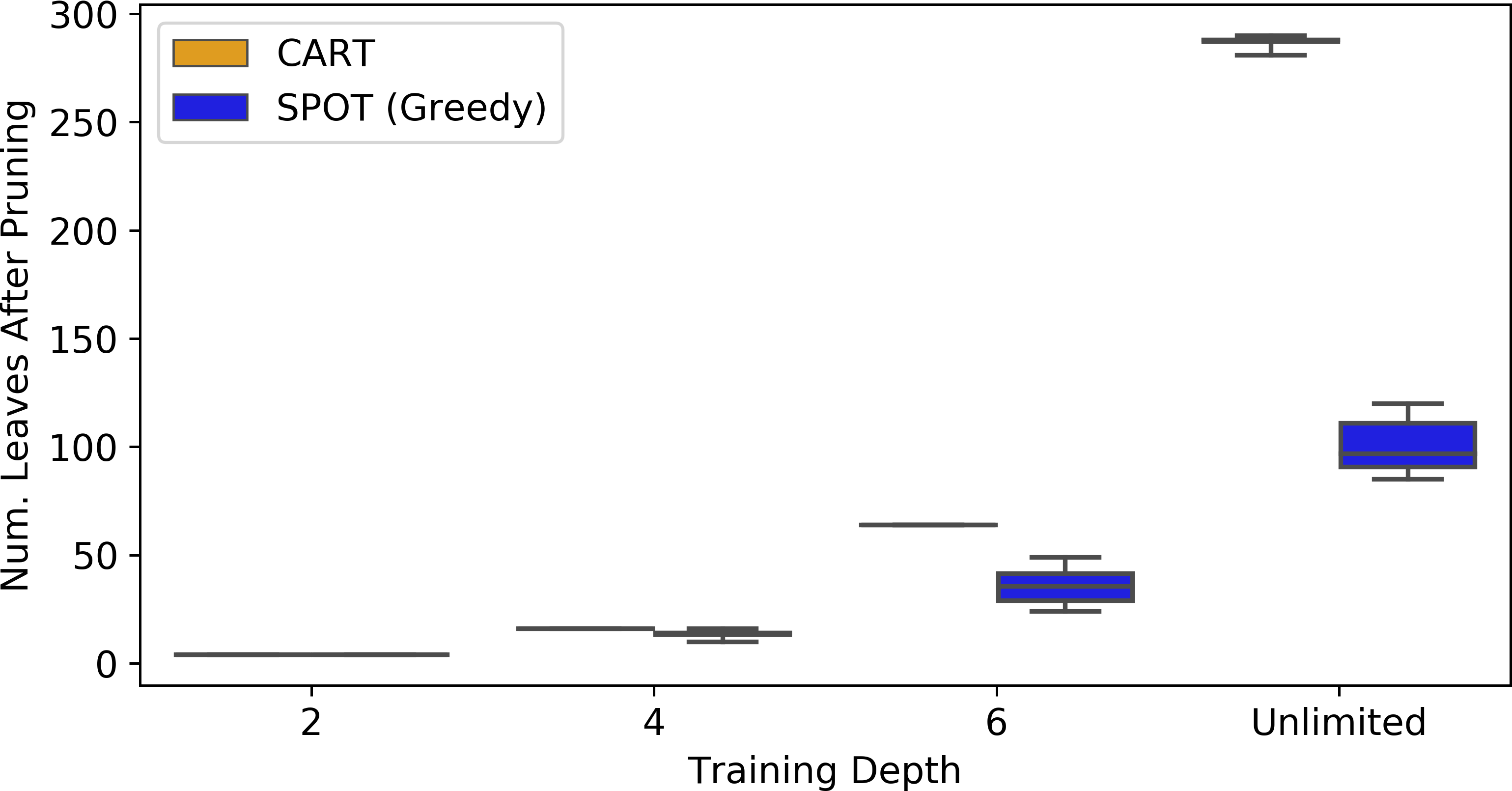}
		\caption{\small $deg = 2$, $\bar{\varepsilon} = 0$}
		\vspace{0.1cm}
		\label{fig:tr10000_eps0_deg2_leaves}
	\end{subfigure}%
	\begin{subfigure}{0.47\linewidth}
		\centering
		\includegraphics[width=0.97\linewidth]{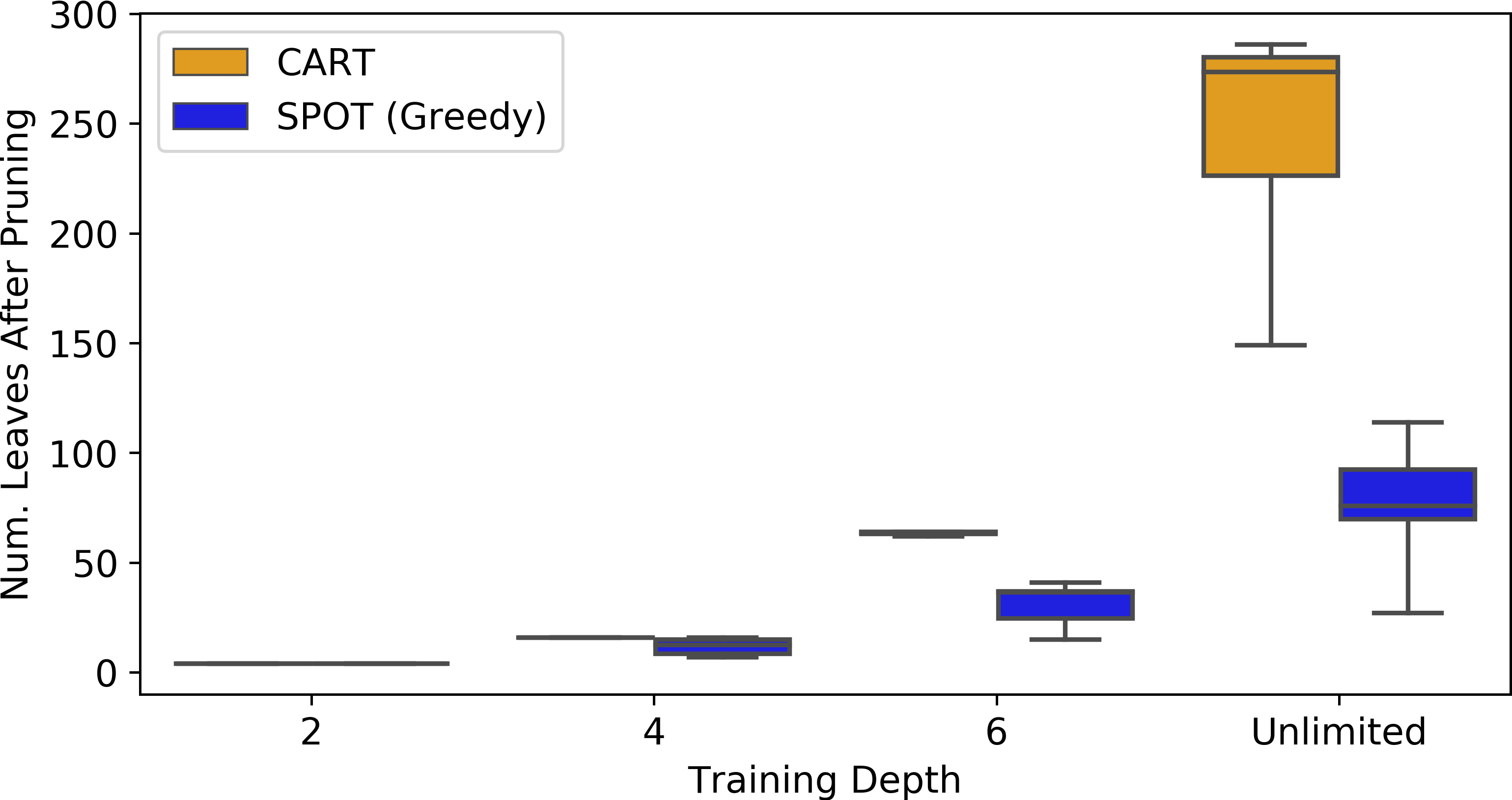}
		\caption{\small $deg = 2$, $\bar{\varepsilon} = 0.5$}
		\vspace{0.1cm}
		\label{fig:tr10000_eps0.5_deg2_leaves}
	\end{subfigure}
	\\
	\begin{subfigure}{0.47\linewidth}
		\centering
		\includegraphics[width=0.97\linewidth]{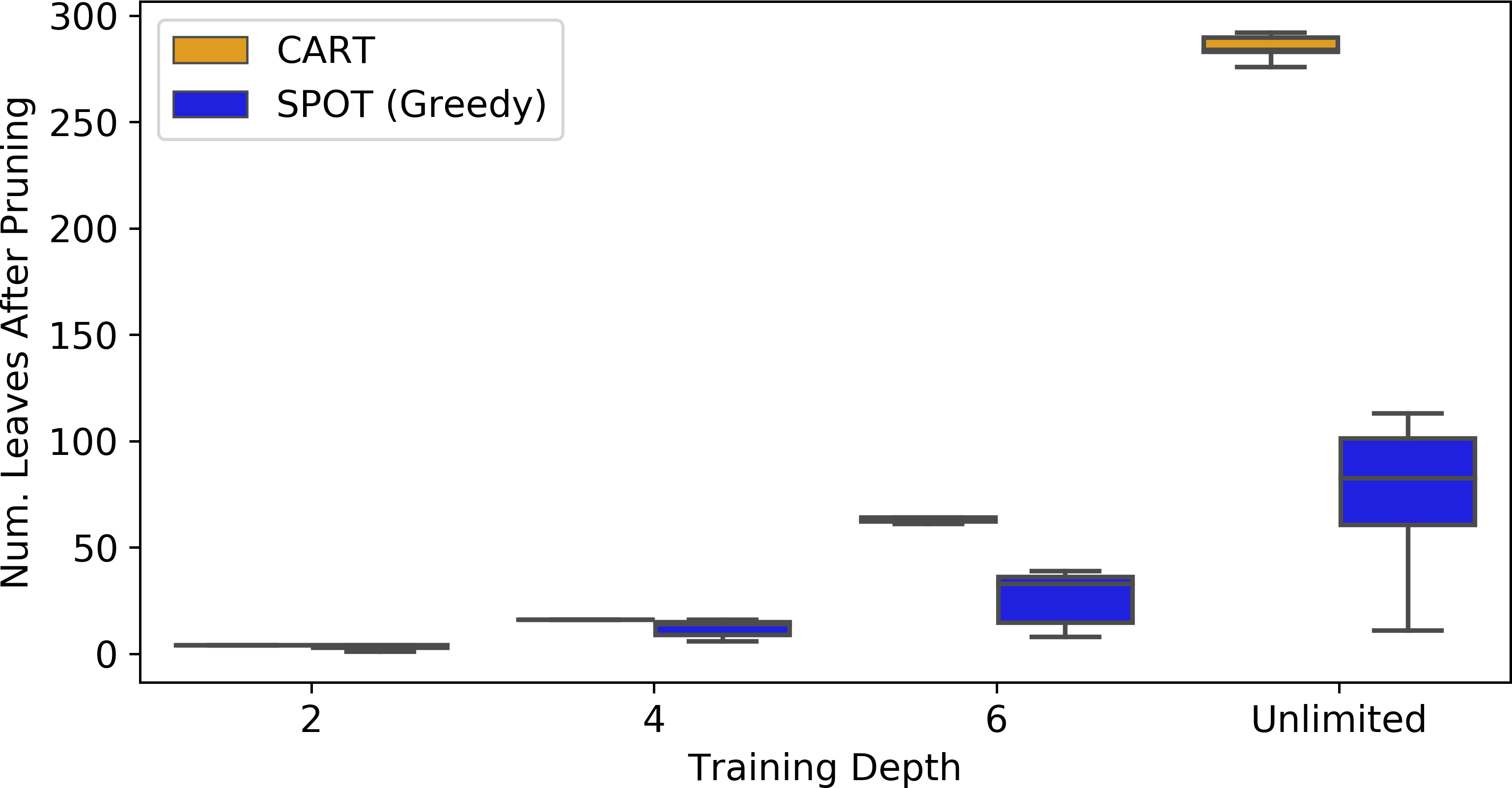}
		\caption{\small $deg = 10$, $\bar{\varepsilon} = 0$}
		\vspace{0.1cm}
		\label{fig:tr10000_eps0_deg10_leaves}
	\end{subfigure}%
	\begin{subfigure}{0.47\linewidth}
		\centering
		\includegraphics[width=0.97\linewidth]{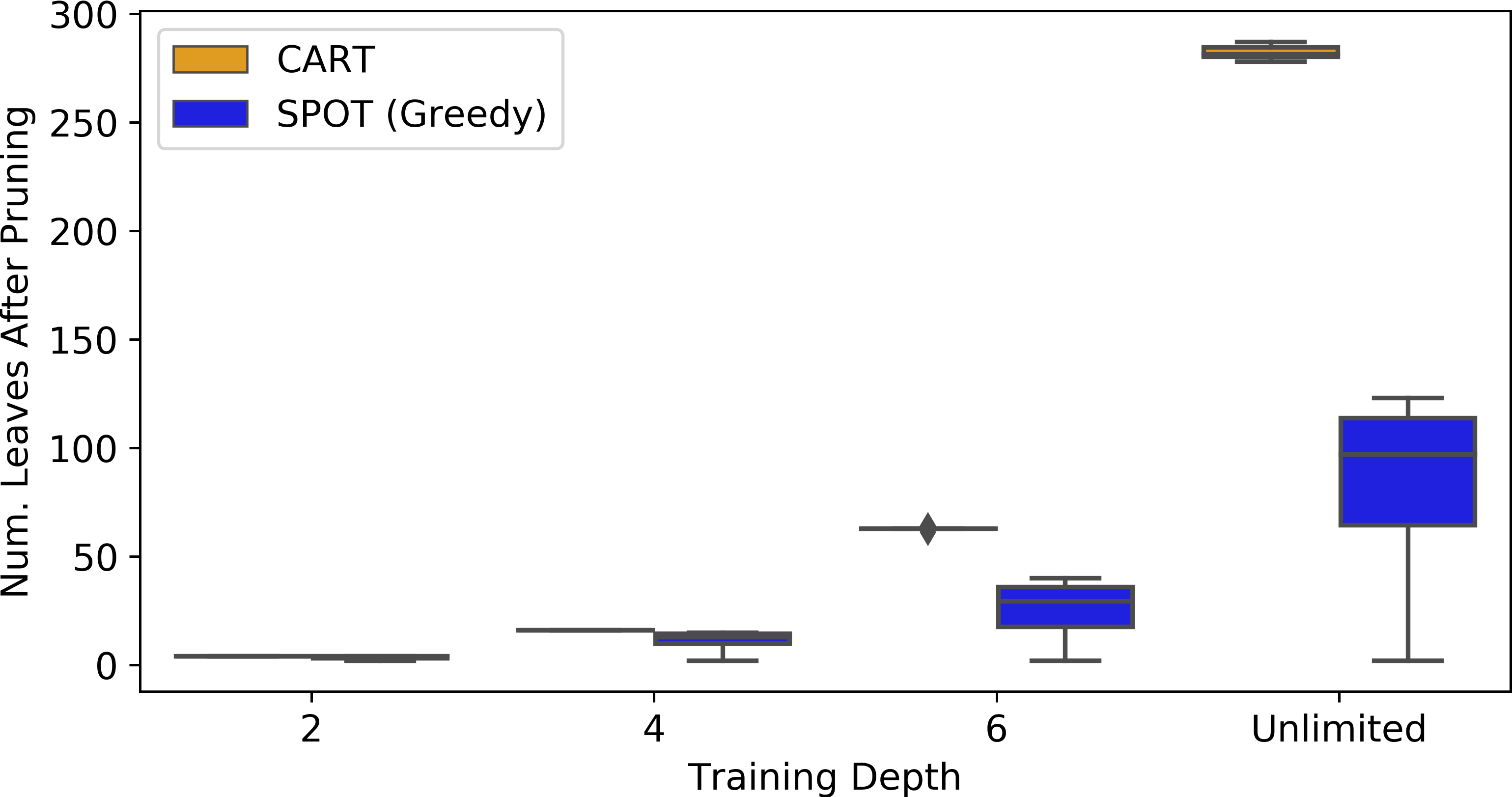}
		\caption{\small $deg = 10$, $\bar{\varepsilon} = 0.5$}
		\vspace{0.1cm}
		\label{fig:tr10000_eps0.5_deg10_leaves}
	\end{subfigure}
	\caption{Number of leaves contained within the SPOT and CART trees from Figure \ref{fig:tr10000}. Each boxplot visualizes the number of leaves associated with the trained trees from 10 different shortest path datasets of size $n = 10000$.}
	\label{fig:tr10000_leaves}
\end{figure}

We also investigate the decision performance of the algorithms on the shortest path problem when trained on larger datasets of $n = 10000$ observations. Since there are more training observations available, it is now feasible to train the decision tree algorithms to higher depths than in the previous experiment. Therefore, we train and evaluate the algorithms on depth sizes up to 6, and we also report the performance of SPOT and CART when trained without any depth restrictions. We also increase the level of noise from $\bar{\varepsilon} = 0.25$ to $\bar{\varepsilon} = 0.5$ to make the estimation problem more challenging for the algorithms given the increased amount of data.

The test set normalized extra travel times incurred by the algorithms for $n=10000$ are given in Figure \ref{fig:tr10000}. As in the previous set of experiments, we observe that the SPO Trees achieve stronger empirical performance over CART when the training depths are restricted to small or modest values, with SPOTs attaining both better average performance and lower variance in performance across the 10 experimental trials. However, when the training depths increase to six or more, CART begins to achieve comparable performance to SPOT and even slightly outperforms SPOT in some cases. Although individual CART splits have little value for decision-making, in combination they finely partition the feature space to a sufficient degree that the predicted cost vectors are highly accurate within each of the resulting leaves. Therefore, CART is eventually able to achieve highly accurate predictions -- and therefore near-optimal decisions -- as its depth increases. However, its interpretabilty is sacrificed as a result, as the trees eventually grow to a size which is too large to be easily visualized and interpreted. 

Figure \ref{fig:tr10000_leaves} reports the number of leaves contained within the learned CART and SPOT trees as a function of their training depths. As the figure demonstrates, when the training depths of CART and SPOT are large or unrestricted, the SPO Trees contain less than half the number of leaf nodes as CART. Therefore, SPO Trees achieve comparable accuracy to CART in these settings \textit{while also} being more concise and therefore more interpretable. We find that the random forest algorithms achieve similar performance, with CART random forests having a very slight edge over SPO Forests in the normalized extra travel times observed on the test set. The greedy SPOT approach also appears to perform similarly to the MILP approach.

\subsection{News Article Recommendation:} We also examine the performance of the SPO Trees and benchmark algorithms on a real dataset. In particular, we consider a news article recommendation problem constructed from the publicly-available Yahoo!\@ Front Page Today Module dataset \citep{web:yahoo}. In the problem we construct, a news aggregation service recommends an article belonging to one of $d$ article types to arriving users with the objective of maximizing the probability of each user clicking on the recommended article. User click probabilities for different article types are unknown to the news aggregator but can be estimated using contextual features that characterize user preferences. Given article click probability estimates $p \in \mathbb{R}^{d}$ for an individual user (i.e., the ``costs'' $c$ for this decision problem), the news aggregator solves the following article recommendation problem:
 \begin{align*}
 \label{eqn:newsrecom}
 z^*(p) = \underset{\substack{w\geq 0\\e^T w=1}}{\max}  p^T w ~~~ \text{s.t. } a_m^T w \leq b_m, ~~~  \forall m\in \{1\dots M\}\,,
\end{align*}
where $w_k$ represents the probability that the news aggregator recommends article $k$ to the user for $k \in \{1,...,d\}$, and $a_m \in \mathbb{R}^{d}, b_m \in \mathbb{R}$ for $m \in \{1\dots M\}$ are the corresponding constraints represent certain restrictions on article recommendations (e.g. ensuring that all article types have some non-zero probability of being recommended). The restrictions could naturally involve budgetary constraints -- for example, Facebook intends to pay certain news publishers as much as \$3 million per year to display their news headlines and article previews to visiting users  \citep{web:wst}. 

The Yahoo!\@ Front Page dataset contains 45,811,883 interaction records between users and news articles from May 1, 2009 to May 10, 2009. We used records from May 1-5 for training data and from May 6-10 as test data; 50\% of the training set records were additionally held out to construct a validation set for parameter tuning. The users and displayed articles are each characterized by five continuous features, which were constructed using a conjoint analysis with a bilinear model; see \citet{chu2009case} for more details. We clustered the articles into $d=6$ categories, and we clustered the historical users into 10000 clusters. Each user cluster was used to construct a feature-cost pair $(x,p)$ for the predict-then-optimize problem, in which we (1) computed the average user feature vector for that cluster ($x$), and (2) computed the average click probability for each article type within that cluster ($p$). After filtering out clusters with an insufficient number of interaction records, we were left with 5130, 5105, and 8768 feature-cost pairs in the training, validation, and test sets, respectively. We also define \textit{sample weights} for the feature-cost pairs as the number of interaction records associated with each pair, and we utilize these sample weights in training and testing the algorithms. The full details of our preprocessing methodology are given in Appendix \ref{sec:expdetsnr}.

\begin{figure}
	\centering
	\includegraphics[width=0.6\linewidth]{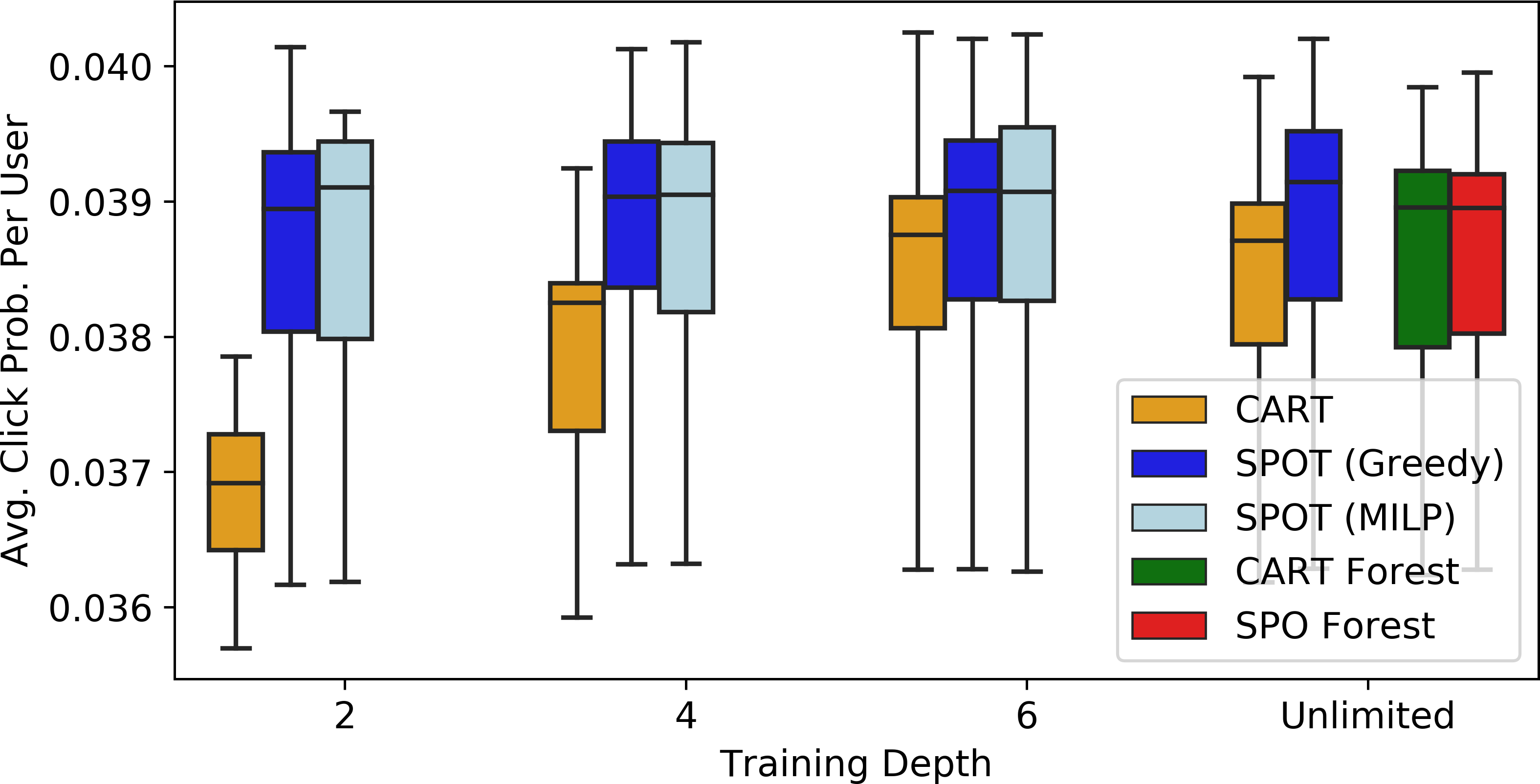}
	\caption{Test set average click probabilities on 9 different constraint sets.}
	\label{fig:newsres}
\end{figure}

The tree and forest algorithms are trained using a minimum leaf size of 10000 interaction records (computed using the sample weights), and the SPOT and CART algorithms are additionally pruned using the held-out validation set. The forest algorithms are trained using $B = 50$ trees with no depth limit, and the number of features $f \in \{2,3,4,5\}$ to use in feature bagging is tuned on the validation set. The empirical runtimes of our algorithms are discussed in Appendix \ref{sec:expdetsnr}. 
We generate $M = 5$ decision feasibility constraints by sampling each element of $a_m$ from an $Exponential(1)$ distribution and setting $b_m = 1$ for $m \in \{1,...,5\}$. Figure \ref{fig:newsres} visualizes the test set performance of the algorithms on 9 different constraint sets generated using the above procedure. Test set performance is defined as the average test set click probabilities of an algorithm's recommended articles, where the average is weighted over test set instances according to the sample weights (equivalent to measuring SPO loss). 
As in the previous section, we find that SPO Trees of very shallow depth outperform CART trees of \textit{unrestricted} depth. Specifically, a greedy SPO Tree of depth 2 achieves percentage improvements in average click probability of 4.3\%, 1.6\%, 0.05\%, and 0.17\% over CART trained to depths of 2, 4, 6, and unrestricted depth, respectively. The MILP SPOT approach appears to perform similarly to the greedy approach. The CART Forest and SPO Forest methods also perform similarly, but surprisingly achieve slightly lower click probabilities than an individual SPO Tree, which may be due to the forest methods overfitting on the training set.

\pdfoutput=1
\section{Conclusion}
We propose tractable methodologies for training decision trees under SPO loss within the predict-then-optimize framework. Our results demonstrate that SPOTs capably produce trees that simultaneously provide higher quality decisions and lower model complexity than de facto tree-building methods designed to minimize prediction error.

\subsection*{Acknowledgments}
Elmachtoub and McNellis were partially supported by NSF grant CMMI-1763000.
\bibliographystyle{informs2014}
\bibliography{ms.bib}
\newpage
\pdfoutput=1
\begin{center}
\vspace{0.8cm}
    \Large Appendices for\\
    \vspace{0.2cm}
    \Large \textbf{Decision Trees for Decision-Making under the \\ Predict-then-Optimize Framework}
    \noindent\makebox[\linewidth]{\rule{0.8\paperwidth}{0.9pt}}
\end{center}

\begin{APPENDICES}

\section{Encoding Decision Trees using Integer and Linear Constraints} \label{sec:milpdet}

Here we provide the complete formulation of $r_{il} \in \mathcal{T}$ as integer and linear constraints using the decision tree encoding proposed in \citet{bertsimas2017optimal}. As it is only covered briefly here, we encourage the reader to examine \citet{bertsimas2017optimal} for a more thorough treatment of the materials below. We assume that the practitioner has specified the following parameters regulating the growth of the tree during the training procedure: (1) the depth $H$ of the tree being trained, and (2) the minimum number of training observations $N_{min}$ permitted to be in each leaf of the tree. We consider training a \textit{complete} tree of depth $H$, define as a tree in which all leaves have a depth of $H$. Let $L$ denote the number of leaves in the tree, and index each leaf by $l \in \mathcal{T}_L := \{1,2,...,L\}$. Further, let $B$ denote the number of branch nodes (i.e., splitting nodes) within the tree, and index each branch node by $t \in \mathcal{T}_B := \{1,2,...,B\}$. Note that $L = 2^H$ and $B = 2^H - 1$. Not all leaves in the tree are required to be active (i.e., contain training observations), and not all branch nodes are required to be active splits (i.e., partition the training observations). Indeed, leaves may be pooled together if their parent splits do not contribute significantly to minimizing the objective function. To keep track of the active leaves and branch nodes, let $k_{l}$= $\mathbb{I}$\{leaf $l$ is not empty$\}$ and $d_{t}$= $\mathbb{I}$\{branch node $t$ is an active split$\}$. If a branch node is \textit{not} an active split, then it effectively considered as a leaf with respect to the complete tree by (1) having all observations take the path corresponding to its left branch, and (2) constraining all child branch nodes to also not be active splits.

We assume without loss of generality that all feature components are numeric and belong to the interval $[0,1]$. Note that categorical features can be easily transformed to fit this assumption through binarization. Each decision tree split is encoded through the variables $a_t \in \{0,1\}^p$ and $b_t \in [0,1]$. The variable $a_t$ indicates which feature component is involved with the split, and $b_t$ indicates the splitting point. For example, if there are three feature components, then the split $``x_2 < 0.4''$ is encoded by $a^T x < b$ where $a = [0,1,0]$ and $b = 0.4$. Since decision tree splits only consider one feature component at a time, only one entry of $a_t$ is permitted to be nonzero. Note that the quantities $a_t$ and $b_t$ are treated as additional decision variables in the SPO Tree MILP as well as $k_l$ and $d_t$.

Let $p(t)$ denote the parent node of $t$. Further, let $A_L(t)$ be the set of left ancestor nodes of node $t$, defined as the set of ancestors of $t$ whose left branch has been followed on the path from the root node to $t$. Define $A_R(t)$ similarly as the set of right ancestor nodes of $t$. 

The constraint $r_{il} \in \mathcal{T}$ in the SPO Tree MILP may be replaced with the set of linear and integer constraints below developed by \citet{bertsimas2017optimal} to encode the splitting logic of decision trees:
\begin{subequations}
\begin{align}
	& \sum_{l=1}^L r_{il}= 1 ,\quad \forall i\in  \{1,2,...,n\} \\
	& r_{il}\leq k_l,\quad \forall i\in  \{1,2,...,n\}, l\in  \mathcal{T}_L \\
	& \sum_{i = 1}^n r_{il}\geq N_{min}k_l,\quad \forall l\in  \mathcal{T}_L \\
    &  a_m^T x_i \geq b_m - (1-r_{il}),\quad \forall l\in  \mathcal{T}_L, i\in  \{1,2,...,n\},  m\in A_R(l) \label{eqn:split1}\\
    & a_m^T (x_i + \epsilon)\leq b_m  + (1+\epsilon_{max}) (1-r_{il}),\quad \forall l\in  \mathcal{T}_L, i\in  \{1,2,...,n\}, m\in A_L(l) \label{eqn:split2}\\
    & \sum_{j = 1}^p a_{jt}= d_t,\quad \forall t \in T_B\\
    & 1-d_t \leq b_t \leq 1,\quad \forall t\in T_B \label{eqn:correction} \\ 
    & d_t \leq d_{p(t)},\quad \forall t\in T_B / \{1\} \\
    & a_{jt}, d_t \in \{0,1\}, \quad \forall j\in \{1...p\}, t \in T_B\\
    & r_{il}, k_l \in \{0,1\} , \quad \forall i\in \{1...n \}, l\in  \mathcal{T}_L
\end{align}
\end{subequations}

Above, $\epsilon_j = \left\{ x_j^{(q+1)} - x_j^{(q)} |  x_j^{(q+1)} \neq x_j^{(q)} \quad 1 = 1, 2, \dots n-1 \right\}$ is the smallest nonzero difference between observed values of feature component $j$, where $x_j^{(q)}$ is the $q^{th}$ largest value observed for feature $x_j$ and $\epsilon_{max} = \max_j {\epsilon_j}$. We encourage the reader to consult \citet{bertsimas2017optimal} for intuition regarding $\epsilon$ and its role in the constraints.

In \citet{bertsimas2017optimal}, if a branch node is considered to be inactive, then its associated split parameters $a$ and $b$ are set to the zero vector and zero, respectively. This design choice was intended by the authors to force all training observations down the right branch by making the left split direction constraint (\ref{eqn:split2}) infeasible for all training observations. However, we believe that this logic was implemented incorrectly, as \textit{both} constraints (\ref{eqn:split1}) and (\ref{eqn:split2}) are feasible for any training observations when $a$ and $b$ are both zero. We have corrected for this behavior by modifying constraint (\ref{eqn:correction}) to set $b$ equal to \textit{one} when a branch node is inactive, therefore successfully making constraint (\ref{eqn:split1}) infeasible when $a$ is the zero vector and forcing observations down the left branch. 

\section{SPOT Integer Programming Approach: Additional Implementation Details} \label{sec:milpdet2}

To prevent unnecessarily large trees and overfitting, \citet{bertsimas2017optimal} recommend adding the quantity ``$\alpha \sum_{t \in T_{B}} d_t$'' to the objective function to penalize trees with a large number of active splits. The parameter $\alpha$ is intended to be chosen by the practitioner to balance the trade-off between concise trees and low training set error, and this parameter can be tuned through applying methods such as cross-validation. However, cross-validation might not be feasible in situations where solving the optimization problem is too computationally expensive to be performed for multiple values of $\alpha$ across multiple folds. In our numerical experiments, we train the SPO Trees with no regularization and instead apply the well-known CART post-pruning algorithm (using SPO loss) proposed by \citet{breiman1984classification} to regularize the tree. To avoid lengthy technical details, we refer the reader to \citet{breiman1984classification} for more information about the pruning algorithm.

Finally, we detail a few strategies for improving the computational time associated with solving the mixed integer linear program. First, as noted in Section \ref{sec:ipapproach} of the main paper, we recommend warm starting the MILP with the solution recovered from the greedy algorithm. Second, we have observed that the computational time is influenced by the precision of the vector of constants $\epsilon$. Since the magnitude of $\epsilon$ is tied to the smallest (nonzero) differences between feature values, we recommend rounding the features according to a certain precision (e.g., $1e^{-2}$) in settings where feature rounding would not affect the quality of the resulting decision tree. Finally, we have observed that the linear programming (LP) relaxation of the MILP often has large negative solutions, which can slow down MILP solvers which rely on LP relaxations to bound the objective function (e.g., branch and bound). We recommend including the following constraint to ensure that the LP relaxation associated with the MILP has at least a lower objective function bound of zero:
\begin{align*}
    \left( \sum_{l = 1}^L y_{il} \right) - z^*(c_i) \geq 0 \quad \forall i \in \{1,2,...,n\}
\end{align*}



\section{Additional Experimental Details: News Article Recommendation}\label{sec:expdetsnr}

First, we provide a more thorough description of how we preprocessed the Yahoo!\@ Front Page Today Module dataset. The dataset contains 45,811,883 interaction records between users and news articles from May 1, 2009 to May 10th, 2009. Each record entry consists of: a feature vector of dimension 5 that characterizes the visiting user, a feature vector of dimension 5 encoding the article displayed to the user, and finally a binary scalar representing whether the user clicked on the displayed article. The user and article features were constructed using a conjoint analysis with a bilinear model; see \citet{chu2009case} for more details. We preprocessed the dataset according to the following procedure in order to obtain training, validation, and test sets of feature-cost pairs for use in our predict-then-optimize problem.
\begin{enumerate}
    \item Randomly sample without replacement $50\%$ of the interaction records from May 1, 2009 to May 5, 2009 for training, and use the rest for validation. The test data consists of all records from May 6, 2009 to May 10, 2009.
    \item Cluster users into 10,000 clusters (``user types'') by applying the $K$-means algorithm to the user features observed in the training and validation data, and similarly cluster the displayed articles into 7 clusters (``article types'') using the article features. For each user cluster, record the mean user feature vector associated with all training and validation set interaction records that map to that cluster.  
    \item Apply the following procedure separately to the training, validation, and test sets of interaction records. For each set of data, group the interaction records according to user type using the clustering obtained in the previous step. Each of these user types corresponds to a feature-cost pair $(x,p)$ for the predict-then-optimize problem. The features $x$ are derived by looking up the mean user feature vector associated with the given cluster computed in the previous step. The costs $p$ are derived by computing the average click probability of each article type across the interaction records associated with  the given cluster. Here, we note that we dropped one article type as well as a number of feature-cost pairs in the training, validation, and test sets to ensure the average click probabilities for each user and article type were calculated with at least 50 interaction records. We were left with 6 article types and 5130, 5105, and 8768 feature-cost pairs in the training, validation, and test sets, respectively.
\end{enumerate}

We also note the empirical runtimes of our algorithms on this dataset. The greedy SPO Trees were trained on a Dell PowerEdge M915 Linux server using 1 processor core and 1 GB of memory per tree. The greedy SPOT training procedure (using unrestricted depth) terminated after at most 1.3 hours for each constraint set, yielding trees of depths between 28 and 38 before pruning (after pruning, the trees had an average depth of 7). SPO Forests were trained on the same server parallelizing fitting trees in the forest across 10 cores and using 40 GBs of memory. The SPO Forests training procedure terminated after at most 18.4 hours of computational time per constraint set.

\end{APPENDICES}

\end{document}